\documentclass[journal]{IEEEtran}

\usepackage[T1]{fontenc}
\usepackage{graphicx}
\usepackage{amsmath,amssymb,amsthm}
\usepackage{booktabs}
\usepackage{array}
\usepackage{multirow}
\usepackage{algorithm}
\usepackage{algpseudocode}
\usepackage{microtype}
\usepackage[dvipsnames]{xcolor}
\usepackage[shortlabels,inline]{enumitem}
\usepackage{cite}
\usepackage[hyphens]{url}

\usepackage[breaklinks,colorlinks,allcolors=black,hypertexnames=false]{hyperref}
\usepackage[capitalize,noabbrev]{cleveref}

\theoremstyle{plain}
\newtheorem{theorem}{Theorem}
\newtheorem{proposition}[theorem]{Proposition}

\theoremstyle{definition}

\theoremstyle{remark}

\crefname{conjecture}{Conjecture}{Conjectures}
\Crefname{conjecture}{Conjecture}{Conjectures}
\crefname{remark}{Remark}{Remarks}
\Crefname{remark}{Remark}{Remarks}

\newcommand{\ED}{\mathrm{ED}}
\newcommand{\EDs}{\mathrm{ED}^{*}}
\newcommand{\xobs}{x_{\mathrm{obs}}}
\newcommand{\nuobs}{\nu_{\mathrm{obs}}}
\newcommand{\numax}{\nu_{\mathrm{max}}}
\newcommand{\muref}{\mu_{\mathrm{ref}}}
\newcommand{\XW}{\mathcal{X}}
\newcommand{\pd}{p_{\mathrm{data}}}
\newcommand{\pmo}{p_{\mathrm{model}}}
\newcommand{\dB}{\,\mathrm{dB}}
\newcommand{\E}{\mathbb{E}}
\newcommand{\Var}{\operatorname{Var}}
\newcommand{\NFE}{\mathrm{NFE}}
\newcommand{\pbridge}{$p$-Bridge}

  {\par\medskip\begingroup\color{red}\noindent\textbf{[TODO]}\itshape\ }%
  {\endgroup\par\medskip}

\renewcommand{\arraystretch}{1.15}

\def\papertitle{The Noise Plane: A Diffusion Bridge Across Exponent and Dispersion}

\newcommand{\maketitlesupplementary}{%
  \onecolumn
  {%
    \centering
    \vspace*{0.5em}
    {\LARGE\bfseries Supplementary Material\par}
    \vspace{0.7em}
    {\large\papertitle\par}
    \vspace{2em}
  }%
}

\begin{document}

\title{\papertitle}

\author{Xuran~Hu,
	Yuru~Jia,
	Tengxi~Wang,
	Milo\v{s}~Dakovi\'{c},~\IEEEmembership{Member,~IEEE,}
	Ljubi\v{s}a~Stankovi\'{c},~\IEEEmembership{Fellow,~IEEE,}
	and~Wufan~Zhao,~\IEEEmembership{Member,~IEEE}
	\thanks{Xuran Hu, Tengxi Wang and Wufan Zhao are with The Hong Kong University of Science and Technology (Guangzhou), Guangzhou 511453, China (e-mail: xuranh@kth.se, twang744@connect.hkust-gz.edu.cn, wufanzhao@hkust-gz.edu.cn).}
	\thanks{Xuran Hu and Yuru Jia are with KTH Royal Institute of Technology, SE-100 44 Stockholm, Sweden (e-mail: yuruj@kth.se).}
	\thanks{Milo\v{s} Dakovi\'{c} and Ljubi\v{s}a Stankovi\'{c} are with University of Montenegro, 81000 Podgorica, Montenegro (e-mail: milos@ucg.ac.me; ljubisa@ucg.ac.me).}
	\thanks{\textit{(Corresponding author: Wufan Zhao.)}}}

\markboth{}%
{}

\maketitle
\begin{abstract}
Diffusion-based denoising commonly uses Gaussian perturbations, whereas sensor noise follows diverse signal-dependent statistical laws. In this paper, we propose $p$-Bridge, a unified measurement-indexed diffusion framework for Tweedie noise. The noise exponent $p$ describes the variance--intensity relation, while accumulated measurement $\nu$ controls the noise level. Together, they define a noise plane that includes Gaussian, Poisson, and Gamma models. The stochastic bridge connects measurement distributions through independent increments, giving denoising a physical interpretation as progressive measurement accumulation. Both the starting and stopping indices can be specified, allowing restoration from an observed measurement level to a prescribed higher level. Measurement superposition yields a shared reverse-mean update, paired with a deviance-based objective for learning the required conditional mean. A latent compound Poisson--Gamma construction enables sampling at intermediate exponents. A single exponent-conditioned network adapts to different Tweedie noise models without separate training for each exponent. Experiments show that exponent conditioning improves cross-exponent robustness over fixed-exponent models. Using estimated exponents, the shared network achieves robustness comparable to an oracle-selected set of five specialist models. Experiments on synthetic images, FMD fluorescence microscopy, and multi-sensor SAR images further demonstrate its effectiveness across different signal-dependent noise regimes.
\end{abstract}


\IEEEpeerreviewmaketitle

\section{Introduction}
\label{sec:intro}

\IEEEPARstart{I}{mage} denoising is a fundamental problem in computational imaging, where the objective is to recover scene information from measurements corrupted during acquisition. Classical methods combine an observation model with image priors based on sparsity or nonlocal similarity~\cite{dabov2007bm3d}. These approaches established effective ways to separate image structure from noise, while subsequent plug-and-play methods incorporated learned denoisers into model-based optimization~\cite{zhang2022dpir}. Their formulation requires both a useful prior for the clean image and an appropriate statistical description of the observation. In physical imaging, however, the observation statistics vary substantially across sensing mechanisms. Photon counting produces signal-dependent fluctuations in cameras and fluorescence microscopy~\cite{foi2008,boulanger2010}, whereas coherent imaging introduces multiplicative speckle~\cite{goodman1976}. Consequently, a noise level alone is insufficient to characterize the degradation across these different imaging regimes.

Learning-based methods have substantially improved denoising by learning image statistics and restoration mappings from data. Convolutional networks provide efficient inference~\cite{zhang2017dncnn,chen2022nafnet}, Transformer-based models capture long-range image dependencies~\cite{zamir2022restormer}, and noise-conditioned models allow one network to handle different corruption levels~\cite{zhang2018ffdnet}. Nevertheless, supervised restoration depends on how well the training data represent the deployment conditions, while collecting accurate clean references can be difficult in real photography and scientific imaging~\cite{abdelhamed2018,zhang2019fmd}. Learning from noisy observations reduces this dependence. Noise2Noise uses corrupted training targets~\cite{lehtinen2018noise2noise}, and Noise2Self exploits noise independence to learn without clean targets~\cite{batson2019noise2self}. These methods broaden the available training settings, but their validity and performance still depend on the statistical structure of the observations. Blind restoration also benefits from explicitly estimating the degradation. VIRNet, for example, jointly estimates noise and restores the image under a spatially varying Gaussian likelihood~\cite{yue2024virnet}. Such developments make denoising more adaptive, while leaving the choice of the underlying noise distribution an important modeling decision.

\begin{figure}[!t]
	\centering
	\includegraphics[width=\columnwidth]{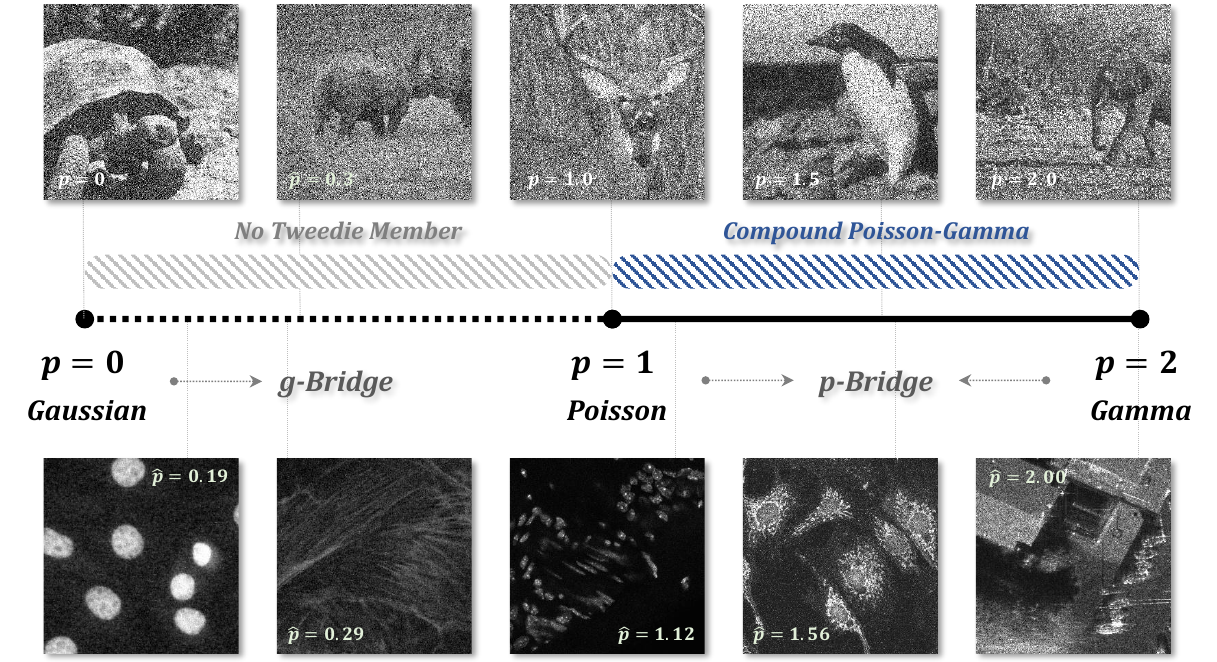}
	\caption{Noise regimes along the Tweedie exponent axis. Top: illustrative noise at representative exponents. Bottom: real observations with estimated exponents $\hat p$.}
	\label{fig:tweedie-axis}
\end{figure}

A complementary line of research addresses this modeling decision through sensor physics and noise transformations. Physics-based noise models describe the acquisition process and generate realistic training pairs~\cite{wei2022physics}, while recent physics-informed approaches learn difficult-to-model sensor components from calibration data such as dark frames~\cite{feng2026pnnp}. These methods improve the connection between synthetic training noise and real measurements, although adapting them to another sensing mechanism requires a corresponding noise model and calibration procedure. Variance-stabilizing transformations offer another route by reducing signal dependence before Gaussian denoising~\cite{anscombe1948,makitalo2013generalized,dalsasso2020sar}. Noise2VST extends this idea by learning a model-free transformation from the noisy input and using a frozen Gaussian denoiser~\cite{herbreteau2025noise2vst}. This demonstrates that a Gaussian denoiser can remain useful for non-Gaussian observations when coupled with suitable adaptation. These approaches focus on modeling or transforming an observation without specifying how its distribution should evolve along a denoising trajectory.

Diffusion-based restoration makes this trajectory an explicit part of the model. Restoration bridges such as I$^2$SB learn processes connecting degraded and clean image distributions~\cite{liu2023i2sb}, while diffusion posterior sampling incorporates measurement likelihoods, including Poisson noise, into reconstruction with a diffusion prior~\cite{chung2023dps}. Beyond Gaussian formulations, Xie et al. construct denoising diffusion processes for Gaussian, Gamma, and Poisson noise with a unified training strategy~\cite{xie2023generative}. Gamma Bridge further gives the trajectory a physical coordinate through the number of looks, allowing restoration between specified measurement levels for Gamma speckle~\cite{gammabridge}. These advances establish both noise-aware diffusion and measurement-indexed restoration, but these constructions treat particular noise laws or discrete families. Extending this perspective to a continuously varying variance--intensity relation requires a common construction that adapts the noise law while preserving the measurement interpretation of the intermediate states.

\begin{figure}[t]
	\centering
	\includegraphics[width=\columnwidth]{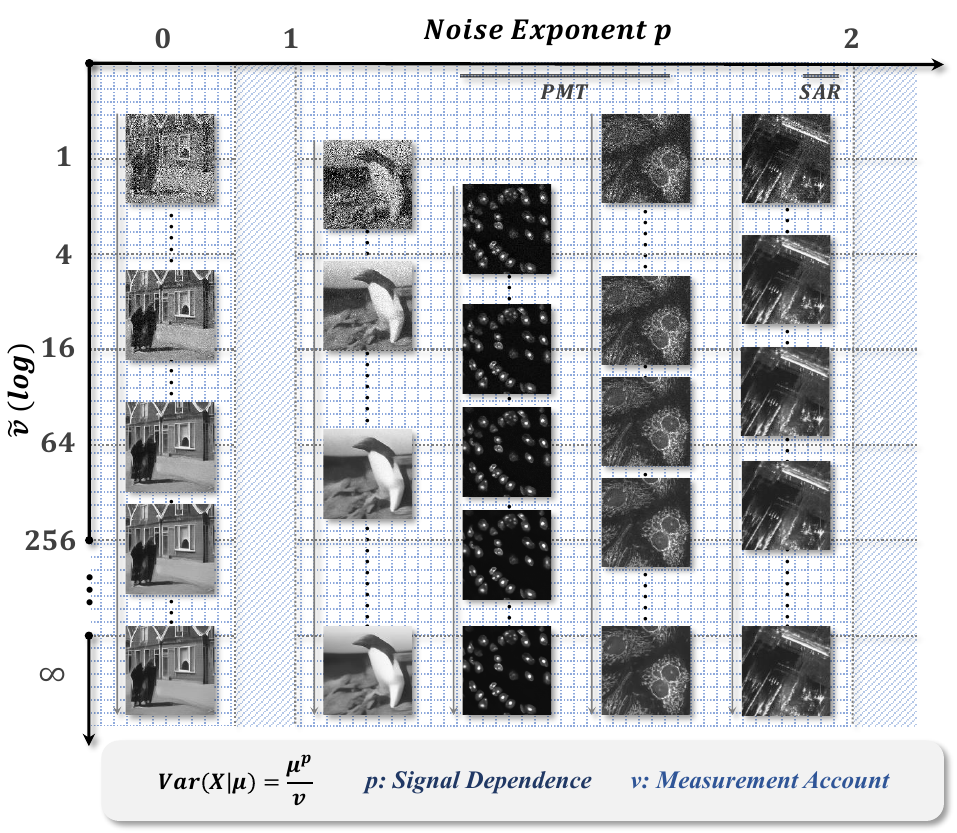}
	\caption{Measurement-indexed denoising trajectories on the noise plane. $p$ controls signal dependence, while increasing $\nu$ reduces noise.}
	\label{fig:noise-plane}
\end{figure}

In this work, we propose \pbridge, a unified measurement-indexed diffusion bridge based on the Tweedie family~\cite{tweedie1984,jorgensen1987}. Our starting point is that accumulating independent measurements of the same scene reduces noise while preserving the expected intensity. The Tweedie variance relation $\mu^p/\nu$ allows us to apply this principle across different noise laws. Here, $p$ describes how variance depends on intensity, while $\nu$ represents accumulated exposure, dose, or number of looks. These two coordinates distinguish changes in the noise law (\cref{fig:tweedie-axis}) from movement along a measurement trajectory (\cref{fig:noise-plane}). At a fixed exponent, increments that are independent conditional on the clean image connect the measurement distributions into a stochastic bridge.

In this construction, the network predicts the clean signal needed for each reverse update, while exponent conditioning allows its prediction to adapt to the observation's noise statistics. We evaluate the role of this adaptation through controlled exponent-mismatch and estimation experiments, and assess restoration quality on synthetic images, FMD fluorescence microscopy, and SAR images from multiple sensors.

The main contributions are as follows:
\begin{enumerate}\itemsep0.1em
\item We propose $p$-Bridge, a unified measurement-indexed diffusion framework for Tweedie noise. Its two noise coordinates have distinct roles. The exponent $p$ specifies the variance--intensity relation, while accumulated measurement $\nu$ controls the noise level. The stochastic bridge connects the corresponding measurement distributions, allowing both the starting and stopping measurement levels to be specified.
\item We derive a shared reverse-mean update from measurement superposition, with a mixing coefficient independent of the noise exponent. A unit-deviance reconstruction objective targets the conditional mean required by this update. A latent compound Poisson--Gamma construction enables sampling of measurement transitions at intermediate exponents.
\item We develop a single exponent-conditioned network that adapts to different Tweedie noise models without separate training for each exponent. At inference, the network uses an exponent estimated from the observation. This shared model achieves cross-exponent robustness comparable to an oracle-selected set of five specialist models.
\end{enumerate}

\section{Related work}
\label{sec:related}

\begin{figure*}[t]
	\centering
	\includegraphics[width=0.8\textwidth]{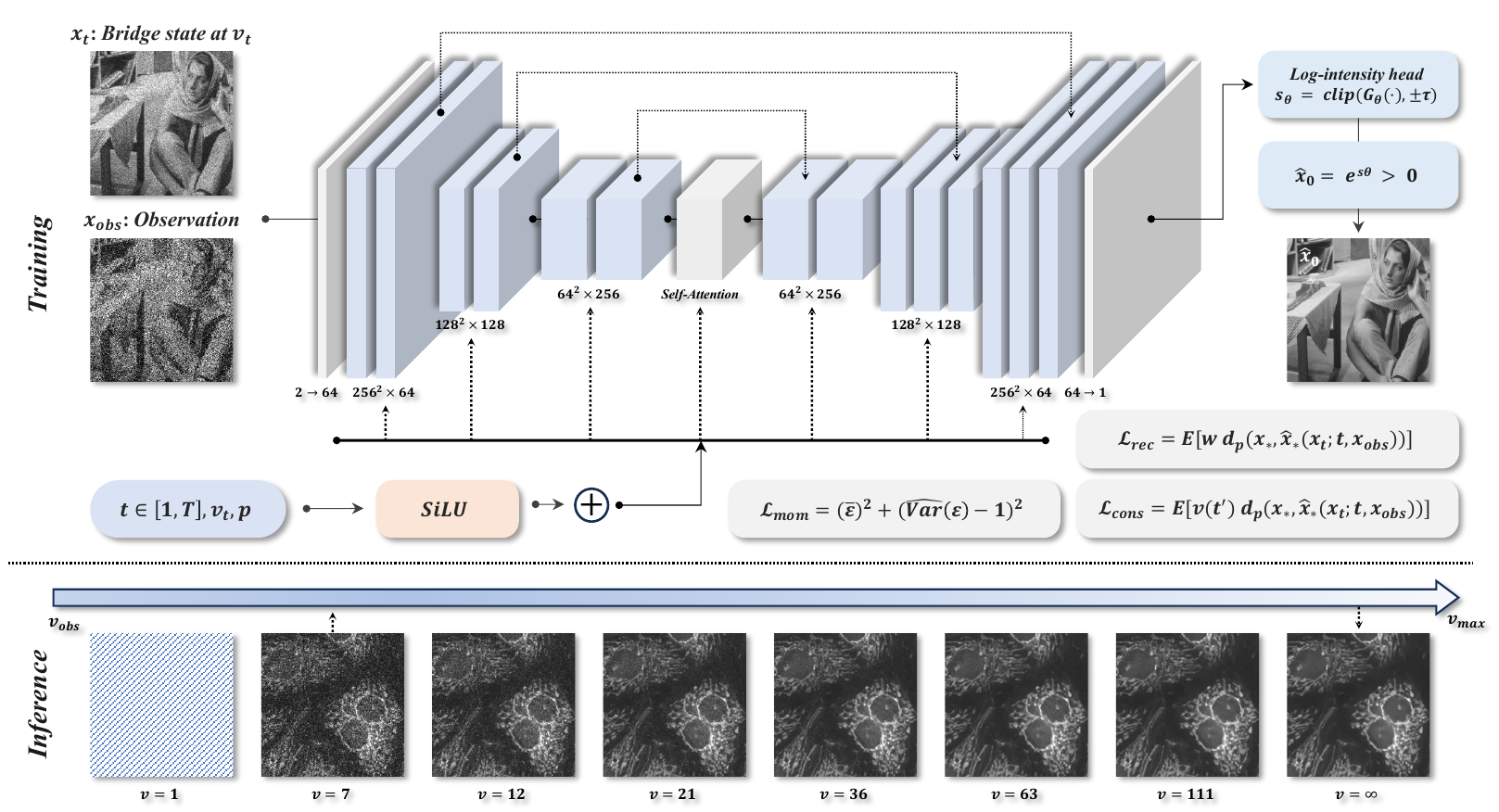}
	\caption{Network architecture and training losses of \pbridge{}, with a deterministic restoration trajectory on an FMD crop.}
	\label{fig:network-overview}
\end{figure*}

\subsection{Diffusion models, bridges and inverse problems}
Widely used denoising diffusion models~\cite{ho2020} and their score-based formulations~\cite{song2021} add Gaussian noise according to a prescribed variance schedule. Subsequent work has improved sampling~\cite{song2021ddim} and network parameterization~\cite{karras2022}. Restoration-oriented methods also redesign the trajectory. For example, ResShift introduces residual shifting to reduce the number of sampling steps~\cite{yue2025resshift}.

Restoration bridges connect clean and degraded image distributions. Schr\"odinger bridge formulations connect prescribed endpoint distributions~\cite{debortoli2021dsb}, with I$^2$SB applying this principle to image restoration~\cite{liu2023i2sb}. Other formulations use direct endpoint interpolation~\cite{delbracio2023} or mean-reverting stochastic differential equations~\cite{luo2023irsde}. More recent developments target spatial adaptation and efficiency. RDBM uses endpoint residuals to modulate noise injection and removal across image regions~\cite{wang2026rdbm}, while E-Bridge shortens the trajectory and learns a consistency-based mapping for single-step restoration~\cite{hou2026ebridge}. System-Embedded Diffusion Bridge Models incorporate the forward system explicitly, embedding a known linear operator and noise covariance into a matrix-valued SDE under an additive Gaussian measurement model~\cite{sobieski2025sdb}.

Another class of methods adapts a pretrained diffusion prior to the observation. DDRM uses the singular-value decomposition of the forward operator~\cite{kawar2022ddrm}, whereas DPS introduces an approximate likelihood score~\cite{chung2023dps}. DPS supports Poisson measurement noise while retaining a Gaussian diffusion prior. IRBridge transfers pretrained generative priors to restoration bridges by relating processes with a common endpoint distribution~\cite{wang2025irbridge}.

Non-Gaussian constructions extend the choice of diffusion distributions. Xie et al. formulate diffusion denoising for Gaussian, Gamma, and Poisson corruption with a unified training strategy~\cite{xie2023generative}. Learning to Jump uses count thinning and reverse count thickening for generative modeling~\cite{chen2023jump}, while Star-Shaped DDPM develops non-Gaussian models within exponential families~\cite{okhotin2023star}. In statistical inference, data thinning splits observations from convolution-closed distributions, including Gaussian, Poisson, and Gamma laws, into independent components~\cite{neufeld2024thinning}. Gamma Bridge indexes a speckle-denoising trajectory by the number of looks~\cite{gammabridge}. \pbridge{} extends this measurement-indexed formulation to the Gaussian point and the Poisson-to-Gamma continuum, using a common superposition construction and a single exponent-conditioned network.

\subsection{Signal-dependent denoising and noise estimation}
Variance-stabilizing transformations reduce signal dependence before Gaussian denoising. Representative choices include the Anscombe transform for Poisson data~\cite{anscombe1948}, with a corrected inverse at low counts~\cite{makitalo2011}, and the logarithmic transform for speckle~\cite{dalsasso2020sar}. Noise2VST learns a model-free transform from a single noisy image and combines it with frozen Gaussian denoisers~\cite{herbreteau2025noise2vst}.

Sensor-specific methods incorporate the statistics of the acquisition process. Photon-limited methods use variance-stabilized nonlocal regression for microscopy~\cite{boulanger2010}, raw-data networks~\cite{chen2018sid}, or physics-based sensor noise models~\cite{wei2020eld}. Speckle methods exploit its multiplicative statistics~\cite{goodman1976} through self-supervised networks~\cite{dalsasso2022} or closed-form shrinkage~\cite{hu2026nonlocal}.

Single-image noise estimation is often formulated within a specified noise family. Gaussian noise estimators use patch statistics and covariance eigenvalues~\cite{liu2013}. Poisson--Gaussian fitting uses local mean--variance relations to estimate shot-noise and read-noise parameters from a raw frame~\cite{foi2008}. VIRNet jointly estimates degradation and restores the image under a spatially varying Gaussian likelihood~\cite{yue2024virnet}. Blind2Sound couples noise-level sensing with self-supervised denoising for Poisson--Gaussian corruption~\cite{liu2025blind2sound}. In \pbridge, the estimated exponent $p$ selects the bridge's measurement law, while the estimated index $\nu$ determines its starting noise level. The two parameters are fitted jointly as the scaling of $\nu$ depends on $p$.

\section{A unified measurement-indexed bridge}
\label{sec:method}

In \pbridge{}, denoising is modeled as the accumulation of independent measurements of the same scene, which reduces noise while preserving the underlying signal. We use the accumulated measurement amount to index a bridge across Gaussian, Poisson, Gamma, and intermediate noise models. The exponent $p$ selects the noise model, while the measurement index $\nu$ controls its noise level. We first construct this bridge, then derive its reverse update and reconstruction loss, and finally describe sampling and training. \Cref{fig:network-overview} shows the network pipelines.

\subsection{Preliminaries and notation}
\label{sec:prelim}

We first introduce the distributional notation used to describe measurements. Let $\mu$ denote a generic mean intensity and $\nu>0$ the accumulated measurement index. The accumulated signal $Z$ follows an additive exponential dispersion model~\cite{jorgensen1987}, written $Z\sim\EDs(\theta,\nu)$, with density
\begin{equation}
	f_Z(z;\theta,\nu)
	=c^{*}(z;\nu)\exp\{\theta z-\nu\kappa(\theta)\}.
	\label{eq:ed-density}
\end{equation}
Here $\theta$ is the natural parameter, $\kappa$ is the cumulant function, and $c^{*}$ is a nonnegative factor independent of $\theta$. The density is taken with respect to a suitable dominating measure, so that continuous, discrete, and mixed distributions are all covered. Normalization makes $\nu\kappa(\theta)$ the log-normalizer. Its first two derivatives give
\begin{equation}
	\E[Z]=\nu\kappa'(\theta),
	\qquad
	\Var(Z)=\nu\kappa''(\theta).
	\label{eq:ed-moments}
\end{equation}

The measured intensity is $x=Z/\nu$. Its mean and variance are determined by the mean map $\tau$ and variance function $V$,
\begin{equation}
	\mu=\tau(\theta)=\kappa'(\theta),
	\qquad
	V(\mu)=\kappa''\!\left(\tau^{-1}(\mu)\right).
	\label{eq:ed-maps}
\end{equation}
Consequently, the intensity has mean $\mu$ and variance $V(\mu)/\nu$. For a fixed noise family, neither $\tau$ nor $V$ depends on $\nu$. Changing the measurement index therefore changes the noise level without changing the underlying mean. Since $\tau'(\theta)=\kappa''(\theta)>0$ in the interior of the natural-parameter domain, $\tau$ is invertible onto the mean domain $\Omega$. Its inverse maps an intensity to the corresponding natural parameter and is the canonical link in generalized linear models.

Tweedie models have the power variance function $V(\mu)=\mu^p$~\cite{tweedie1984,jorgensen1987}. The exponent $p$ specifies the variance--intensity relation, while $\mu$ and $\nu$ determine the mean intensity and measurement level. For the Gaussian branch, we also allow a fixed variance scale $\sigma^2$; setting $\sigma^2=1$ recovers $V(\mu)=\mu^0$. The family-specific functions are listed in \cref{tab:ed-members} of the supplement.

The index $\nu$ represents accumulated measurement rather than noise amplitude. Larger indices correspond to less noisy measurements of the same scene. Accordingly, the forward process decreases $\nu$, whereas denoising increases it. A finite index still describes a noisy measurement; the clean signal is approached as $\nu\to\infty$.

\subsection{Measurement-indexed formulation}
\label{sec:framework}

Let $x_{\star}$ denote the clean image and $x_\nu$ an intensity measurement at index $\nu>0$. Depending on the sensor, $\nu$ represents exposure, dose, or number of looks, expressed in the units of the noise model. We assume
\begin{equation}
  \E[x_\nu\mid x_{\star}]=x_{\star},
  \qquad
  \Var(x_\nu\mid x_{\star})=\frac{V(x_{\star})}{\nu}.
  \label{eq:fwd}
\end{equation}
These moments follow from \cref{eq:ed-moments,eq:ed-maps} with $\mu=x_{\star}$. Any additional dispersion scale is absorbed into $\nu$.

The bridge also requires a rule for combining measurements. For two measurements of the same scene that are independent conditional on $x_{\star}$, their average weighted by $\nu_1$ and $\nu_2$ must remain in the same family at index $\nu_1+\nu_2$. The Tweedie models above have this property. It is simplest to express it using the accumulated signal $Z_\nu=\nu x_\nu$. Write its distribution as $\EDs(\theta,\nu)$, where the natural parameter $\theta$ specifies the underlying mean. For independent $Z_1\sim\EDs(\theta,\nu_1)$ and $Z_2\sim\EDs(\theta,\nu_2)$,
\begin{equation}
  Z_1+Z_2\sim\EDs(\theta,\nu_1+\nu_2).
  \label{eq:add}
\end{equation}
Dividing the sum by $\nu_1+\nu_2$ gives the required intensity average. This includes Poisson superposition and Gamma addition at a common rate. Closure is not guaranteed for different natural parameters.

We now arrange these measurement distributions along a decreasing schedule $\nu(0)=\numax>\cdots>\nu(T-1)=\nuobs$. At each step, define
\begin{equation}
  Z_t\mid x_{\star}\sim\EDs(\theta_{\star},\nu(t)),
  \qquad x_t=\frac{Z_t}{\nu(t)},
  \label{eq:bridge}
\end{equation}
where $\theta_{\star}=\tau^{-1}(x_{\star})$ is the natural parameter corresponding to the clean image. The final marginal matches the observation model at $\nuobs$, whereas the finite initial state $x_0$ still contains noise. The clean image is approached in conditional mean square as $\nu\to\infty$.

Each marginal in \cref{eq:bridge} follows the assumed sensor model. To connect these marginals into a joint process, we use independent increments with the same underlying mean, as defined next. All identities apply pixelwise. The regularity conditions, including the availability of arbitrary positive measurement increments and the invertibility of $\tau$ on the mean domain $\Omega$, are stated in \cref{sec:sup-reg}.

\subsection{Family-wide properties}
\label{sec:family-results}

\subsubsection{Superposition and the reverse mean}

A reverse step moves from $\nu(t)$ to the larger measurement amount $\nu(t-1)$. By \cref{eq:add}, the missing amount $\Delta_t=\nu(t-1)-\nu(t)>0$ can be supplied by an independent measurement increment. This defines the following coupling.

\begin{theorem}[Superposition coupling]
\label{thm:superposition}
For $t=1,\ldots,T-1$, let $Y_{t\to t-1}\sim\EDs(\theta_{\star},\Delta_t)$ be independent of $Z_t$. Then $Z_t+Y_{t\to t-1}\sim\EDs(\theta_{\star},\nu(t-1))$. Equivalently,
\begin{equation}
  x_{t-1}=\alpha_t x_t+\frac{Y_{t\to t-1}}{\nu(t-1)},
  \qquad \alpha_t=\frac{\nu(t)}{\nu(t-1)}.
  \label{eq:superposition}
\end{equation}
\end{theorem}

The stochastic reverse therefore adds an independent measurement increment. At $p=2$ it recovers the Gamma Bridge update~\cite{gammabridge}, and at $p=1$ it gives the corresponding Poisson construction. Replacing $x_{\star}$ with a network prediction $\hat x_{\star}$ defines a plug-in stochastic transition. Replacing its increment with its conditional mean gives a deterministic update.

\begin{theorem}[$p$-independence of the reverse mean]
\label{thm:reverse}
Given $(x_t,\hat x_{\star})$ with $\hat x_{\star}\in\Omega$, draw
$Y_{t\to t-1}\mid(x_t,\hat x_{\star})\sim\EDs(\tau^{-1}(\hat x_{\star}),\Delta_t)$.
Under this model transition,
\begin{equation}
  \E[x_{t-1}\mid x_t,\hat x_{\star}]
  =\alpha_t x_t+(1-\alpha_t)\hat x_{\star}.
  \label{eq:reverse}
\end{equation}
The coefficient $\alpha_t$ depends only on the measurement schedule, not on $p$.
\end{theorem}

The update is the conditional mean of the plug-in transition defined above, obtained from $\E[Y_{t\to t-1}\mid x_t,\hat x_{\star}]=\Delta_t\hat x_{\star}$. With oracle predictions $\hat x_{\star}=x_{\star}$ and the correct observation-model initialization, independent increments reproduce the joint law of the superposition coupling. A fixed observation instead defines its conditional trajectory.

The deterministic update preserves the oracle mean but not the bridge marginals. For a bridge-distributed input, its conditional variance is $\alpha_t V(x_{\star})/\nu(t-1)$ rather than $V(x_{\star})/\nu(t-1)$. Physical measurement marginals and oracle distributional exactness therefore apply to the stochastic construction, while the deterministic sampler follows its reverse mean (\cref{sec:sup-preserve}).

\begin{table*}[!t]
	\centering\footnotesize
	\caption{Bridge states and thinning kernels across the supported exponent axis.}
	\label{tab:families}
	\setlength{\tabcolsep}{6pt}
	\begin{tabular*}{0.75\textwidth}{@{\extracolsep{\fill}}ccccc@{}}
		\toprule
		$p$ & $x_\nu$ & index & $c^{*}(z;\nu)$ & $q(z_t\mid z)$ \\
		\midrule
		$0$     & $x_{\star}+\sigma\xi/\sqrt{\nu}$   & frames   & $\propto e^{-z^{2}/(2\nu\sigma^{2})}$ & $\mathcal{N}(\alpha_tz,\sigma^{2}\alpha_t\Delta_t)$ \\
		$1$     & $\mathrm{Pois}(\nu x_{\star})/\nu$ & exposure & $\nu^{z}/z!$                           & $\mathrm{Binomial}(z,\alpha_t)$ \\
		$(1,2)$ & $Z_\nu/\nu$                  & exposure & series$^{\ast}$                        & Binom. $\rightarrow$ Beta$^{\ast}$ \\
		$2$     & $\Gamma(\nu,\nu/x_{\star})$        & looks    & $z^{\nu-1}/\Gamma(\nu)$                & $z_t/z\sim\mathrm{Beta}(\nu(t),\Delta_t)$ \\
		\bottomrule
	\end{tabular*}
	\par\smallskip
	\begin{minipage}{0.75\textwidth}\footnotesize\raggedright
		Here $z:=z_{t-1}$, $\alpha_t=\nu(t)/\nu(t-1)$, $\Delta_t=\nu(t-1)-\nu(t)$, and $\xi\sim\mathcal{N}(0,1)$. For $1<p<2$, $Z_\nu$ is compound Poisson--Gamma and the asterisk denotes its latent two-stage construction. The Gamma entry in this table uses the shape--rate parameterization. Here, $x_{\star}$ denotes the clean image.
	\end{minipage}
\end{table*}

\subsubsection{The unit deviance}

For $y,\mu\in\Omega$, the unit deviance equals twice the Bregman divergence generated by the convex conjugate $\psi=\kappa^*$~\cite{banerjee2005}:
\begin{equation}
  d(y,\mu)=2B_\psi(y,\mu).
  \label{eq:bregman}
\end{equation}

\begin{theorem}[Mean-optimal reconstruction]
\label{thm:mean}
Let $Y$ take values in the domain of $d$, with $\E[|Y|]<\infty$, $m=\E[Y]\in\Omega$, and $\E[d(Y,m)]<\infty$. Then
\[
  \arg\min_{\mu\in\Omega}\E[d(Y,\mu)]=m
\]
uniquely. Applied conditionally to the clean image, the optimal reconstruction under the same conditions is
\[
  \E[x_{\star}\mid x_t,\xobs].
\]
\end{theorem}

The Bregman representation and mean-optimality proof are derived in \cref{sec:sup-loss}. Our bounded clean-image range $\XW=[\epsilon,1]$, $\epsilon>0$, satisfies these conditions. The result aligns the reconstruction term with the conditional mean used in \cref{eq:reverse}.

Differentiating the deviance gives a shared gradient for the power-variance family:
\begin{equation}
  \partial_\mu d_p(y,\mu)=-\frac{2(y-\mu)}{\mu^p}.
  \label{eq:dp-grad}
\end{equation}
With $d_p(y,y)=0$, this yields squared error at $p=0$, Poisson deviance at $p=1$, and twice the Itakura--Saito divergence at $p=2$, as well as the intermediate deviance forms.

\subsection{Exponent-specific thinning}
\label{sec:exponent}

Training couplings and measurement-level initialization require a downward transition that removes part of an accumulated measurement. Under the superposition coupling, its density or mass is
\begin{equation}
  q(z_t\mid z_{t-1};\theta_{\star})
  =\frac{c^{*}(z_t;\nu(t))\,c^{*}(z_{t-1}-z_t;\Delta_t)}
        {c^{*}(z_{t-1};\nu(t-1))}.
  \label{eq:thin}
\end{equation}
The exponential factor depends only on the conditioned total $z_{t-1}$ and cancels. Hence the kernel is independent of $\theta_{\star}$. For compound laws, \cref{eq:thin} includes endpoint atoms as well as a continuous component. The corresponding measures are specified in \cref{sec:sup-kernels}.

At $p=0,1,2$, the kernels are Gaussian, Binomial, and scaled Beta, respectively (\cref{tab:families}). For $1<p<2$, the compound Poisson--Gamma representation instead provides a two-stage construction.

\begin{proposition}[Latent thinning for $1<p<2$]
\label{prop:latent}
Let $Z_\nu=\sum_{i=1}^{N_\nu}J_i$, where $N_\nu\sim\mathrm{Pois}(\lambda\nu)$ and the jumps are independent $\Gamma(\alpha,\beta)$ variables in shape--scale parameterization, independent of $N_\nu$. Set
\begin{equation}
\begin{aligned}
  \lambda&=\frac{\mu^{2-p}}{2-p},&
  \alpha&=\frac{2-p}{p-1},\\
  \beta&=(p-1)\mu^{p-1}.&&
\end{aligned}
  \label{eq:cpg-params}
\end{equation}
Then $\E[Z_\nu]=\nu\mu$ and $\Var(Z_\nu)=\nu\mu^p$. Conditional on $N_{t-1}=n$ and $Z_{t-1}=z$, sample $N_t=n_1\sim\mathrm{Binomial}(n,\alpha_t)$. For $0<n_1<n$,
\[
  Z_t/z\sim\mathrm{Beta}(n_1\alpha,(n-n_1)\alpha).
\]
If $n_1=0$, set $Z_t=0$; if $n_1=n>0$, set $Z_t=z$. If $n=0$, both totals are zero and no ratio is formed.
\end{proposition}

This combines Poisson thinning of the counts with Beta--Gamma splitting of the amplitudes. When a simulator retains the count, these steps require no count inference. For an observed total alone, the count must first be sampled from its posterior. The weights are independent of the unknown mean and are given in \cref{sec:sup-latent}. The finite truncation used for real observations makes that implementation approximate. The synthetic severity-transfer experiment uses simulator counts when thinning the conditioning channel, a protocol-specific source of information not available at deployment (\cref{sec:obj-transfer}).

Within the specified compound model, a Gamma-distributed gain $G$ has excess-noise factor $F^2=\E[G^2]/\E[G]^2=1+1/\alpha$, yielding
\begin{equation}
  p=2-\frac{1}{F^2}.
  \label{eq:excess}
\end{equation}
This relation holds under the mean-dependent jump scale and rate in \cref{eq:cpg-params}. At fixed $\mu$ and $\nu$, the Tweedie family approaches Poisson and Gamma as $p$ approaches $1$ and $2$. By contrast, changing only the event rate at fixed gain produces a linear variance--mean relation, but not generally a Poisson distribution.

\subsection{Output parameterization and measurement calibration}
\label{sec:dec-head}

For positive clean intensities, the conditional coefficient of variation is
\begin{equation}
  \mathrm{CV}(x_t\mid x_{\star})
  =\frac{x_{\star}^{p/2-1}}{\sqrt{\nu(t)}}.
  \label{eq:cv}
\end{equation}
It is independent of intensity at $p=2$ and increases towards dark regions for $p<2$. This motivates our residual anchoring at $p=2$ and direct log-intensity prediction for $1\le p<2$. Positive means are enforced with an exponential output map for $p\ge1$, while the Gaussian branch uses an identity map.

We distinguish the mean domain $\Omega$, the normalized clean-image range $\XW=[\epsilon,1]$, and the observation support. The sampler clips predictions $\hat x_{\star}$ to $\XW$, not sampled states. This bounded-range projection is an implementation choice, since positivity of the mean alone imposes no upper bound. In particular, compound observations can equal zero or exceed the clean-image range.

For a specified exponent $p$, define on a homogeneous patch $\hat\nu=V(\bar x)/\hat\sigma^2$. A radiometric rescaling $x\mapsto cx$, $c>0$, gives
\begin{equation}
  \hat\nu\longmapsto c^{p-2}\hat\nu.
  \label{eq:scale}
\end{equation}
Only at $p=2$ is the estimated index invariant to multiplicative gain. At other exponents, the index must be reported relative to a specified intensity scale. This motivates calibrating the intensity scale and jointly estimating $(p,\nu)$.

\subsection{Exponent range and Gaussian special case}
\label{sec:dec-ends}

We formulate \pbridge{} over $\{0\}\cup[1,2]$, covering the Gaussian model and the Poisson-to-Gamma continuum.

At $p=0$, allowing a fixed variance scale $V(\mu)=\sigma^2$ and writing $\sigma_t^2=\sigma^2/\nu(t)$ gives
\begin{equation}
  x_t=x_{\star}+\sigma_t\xi,\qquad \xi\sim\mathcal{N}(0,I).
  \label{eq:gauss}
\end{equation}
These marginals have the VE-SDE / EDM noise parameterization~\cite{song2021,karras2022}. Their deterministic update in \cref{eq:reverse} is nevertheless different from an EDM probability-flow Euler step. The coefficients on the current state are $\sigma_{t-1}^2/\sigma_t^2$ and $\sigma_{t-1}/\sigma_t$, respectively. The unit-variance convention gives squared-error deviance. Retaining $\sigma^2$ scales it by $\sigma^{-2}$. Our conditional network covers $p\in[1,2]$, with $p=0$ evaluated as a separate branch.

Adding independent Gaussian read noise generally leaves the single-exponent family. Moreover, a fixed read variance per exposure contributes $\sigma_{\mathrm r}^2/\nu^2$ to normalized intensity, rather than the $1/\nu$ scaling in \cref{eq:fwd}. Extending the bridge to this setting requires a cumulative-noise model with compatible independent increments. For real data, the fitted exponent summarizes the variance--intensity relation over the calibration range.

\begin{table*}[t]
	\centering
	\scriptsize
	\caption{Minimum PSNR (dB) across tested noise exponents, averaged over three corruption seeds at $\NFE=5$. ``Told'' and ``blind'' use true and estimated exponents, respectively. ``Envelope'' uses oracle-selected specialists. $\Delta$ is the blind model's gain over the best fixed-exponent model. Bold highlights the best fixed-exponent result and the conditional result.}
	\label{tab:cond}
	\setlength{\tabcolsep}{6pt}
	\begin{tabular}{lccccccccc}
		\toprule
		& \multicolumn{5}{c}{fixed specialist, worst case over $\pd$}
		& \multicolumn{3}{c}{conditional} & \multirow{2}{*}{envelope} \\
		\cmidrule(lr){2-6}\cmidrule(lr){7-9}
		severity & $p{=}1$ & $1.25$ & $1.5$ & $1.75$ & $2$
		& told $\pd$ & blind $\hat p$ & $\Delta$ & \\
		\midrule
		severe   & 18.997 & 20.150 & \textbf{21.339} & 20.036 & 18.007 & 22.260 & \textbf{22.260} & $+0.921$ & 22.254 \\
		moderate & 20.921 & 21.899 & \textbf{23.112} & 22.903 & 21.747 & 24.635 & \textbf{24.637} & $+1.525$ & 24.631 \\
		mild     & 24.142 & 25.052 & 26.022 & \textbf{26.336} & 25.551 & \textbf{27.436} & 27.435 & $+1.099$ & 27.431 \\
		\bottomrule
	\end{tabular}
\end{table*}

\begin{table*}[t]
	\centering\scriptsize
	\caption{Synthetic-image denoising results. PSNR (dB) / SSIM are averaged over three corruption seeds. Bold and underline mark the best and second-best results per metric. ``---'' denotes no trained parameters.}
	\label{tab:baselines}
	\setlength{\tabcolsep}{2.2pt}
	\begin{tabular}{lccccccccccc}
		\toprule
		\multirow{2}{*}{method} & \multirow{2}{*}{params (M)} & \multirow{2}{*}{$\NFE$}
		& \multicolumn{3}{c}{severe, $\eta{=}1$}
		& \multicolumn{3}{c}{moderate, $\eta{=}0.5$}
		& \multicolumn{3}{c}{mild, $\eta{=}0.25$} \\
		\cmidrule(lr){4-6}\cmidrule(lr){7-9}\cmidrule(lr){10-12}
		& & & $\pd{=}1$ & $1.5$ & $2$ & $1$ & $1.5$ & $2$ & $1$ & $1.5$ & $2$ \\
		\midrule
		\multicolumn{12}{l}{\itshape Classical, transform-based, and Gaussian-bridge baselines} \\
		BM3D~\cite{dabov2007bm3d} & --- & 1 & 18.63/.493 & 17.59/.479 & 16.87/.467 & 23.19/.632 & 22.70/.622 & 22.08/.609 & 26.58/.763 & 26.21/.747 & 25.63/.724 \\
		VST\,+\,BM3D~\cite{anscombe1948,makitalo2011} & --- & 1 & 21.21/.485 & 20.12/.434 & 19.86/.462 & 23.87/.634 & 23.77/.638 & 23.62/.641 & 26.89/.778 & 26.76/.778 & 26.59/.776 \\
		Noise2VST~\cite{herbreteau2025noise2vst} & 32.64 & 1 & 21.99/.543 & 21.53/.521 & 21.05/.499 & 24.71/.691 & 24.56/.689 & 24.40/.685 & \textbf{27.66}/\underline{.812} & 27.55/.812 & \underline{27.38}/.811 \\
		\pbridge, $p{=}0$ & 15.19 & 5 & 18.67/.407 & 17.51/.374 & 16.97/.357 & 21.39/.541 & 20.04/.503 & 19.13/.480 & 24.85/.701 & 23.52/.666 & 22.42/.642 \\
		\multicolumn{12}{l}{\itshape Supervised backbones} \\
		NAFNet~\cite{chen2022nafnet} & 29.16 & 1 & 22.30/.572 & 22.23/.575 & 22.16/.576 & 24.64/.692 & 24.57/.693 & 24.48/.695 & 27.50/.810 & 27.41/.811 & 27.28/.811 \\
		Restormer~\cite{zamir2022restormer} & 26.11 & 1 & 22.17/.570 & 22.09/.570 & 22.02/.573 & 24.55/.688 & 24.47/.689 & 24.37/.691 & 27.43/.806 & 27.35/.808 & 27.22/.808 \\
		EAMamba~\cite{lin2025eamamba} & 25.32 & 1 & 22.28/\underline{.574} & 22.21/.576 & 22.13/\underline{.578} & 24.66/\underline{.695} & 24.59/.697 & 24.50/.698 & 27.50/.811 & 27.44/.813 & 27.31/.813 \\
		\multicolumn{12}{l}{\itshape Diffusion-based methods} \\
		DDRM~\cite{kawar2022ddrm} & 15.11 & 5 & 20.36/.445 & 19.54/.401 & 18.74/.371 & 22.25/.542 & 21.36/.492 & 20.52/.448 & 23.40/.627 & 23.31/.620 & 23.13/.605 \\
		I$^2$SB~\cite{liu2023i2sb} & 15.11 & 5 & 21.61/.455 & 21.58/.461 & 21.51/.465 & 23.95/.608 & 23.94/.612 & 23.85/.612 & 27.07/.774 & 27.10/.781 & 26.99/.781 \\
		ResShift~\cite{yue2025resshift} & 15.11 & 5 & 21.68/.544 & 21.60/.545 & 21.51/.546 & 24.05/.667 & 23.99/.669 & 23.90/.670 & 26.77/.789 & 26.73/.791 & 26.63/.791 \\
		\multicolumn{12}{l}{\itshape Noise-model estimation methods} \\
		FBI-Denoiser~\cite{byun2021fbi} & 2.49 & 1 & 17.44/.238 & 16.42/.210 & 12.96/.110 & 22.74/.509 & 20.19/.350 & 15.23/.165 & 18.85/.240 & 15.94/.178 & 12.97/.103 \\
		VIRNet~\cite{yue2024virnet} & 10.53 & 1 & \underline{22.32}/.570 & 22.26/.572 & \underline{22.18}/.574 & \underline{24.72}/\underline{.695} & \underline{24.66}/.697 & \underline{24.58}/\underline{.700} & 27.55/\underline{.812} & 27.48/.814 & 27.36/\underline{.814} \\
		\multicolumn{12}{l}{\itshape Ours} \\
		\pbridge, fixed $p{=}1.5$ & 15.19 & 5 & 21.35/.465 & \textbf{22.34}/\textbf{.580} & 21.34/.516 & 23.83/.611 & \textbf{24.73}/\textbf{.704} & 23.11/.627 & 26.99/.771 & \textbf{27.58}/\textbf{.819} & 26.02/.763 \\
		\pbridge, conditional blind $\hat p$ & 15.28 & 5 & \textbf{22.39}/\textbf{.577} & \underline{22.33}/\underline{.579} & \textbf{22.26}/\textbf{.582} & \textbf{24.79}/\textbf{.700} & \textbf{24.73}/\underline{.703} & \textbf{24.64}/\textbf{.704} & \underline{27.64}/\textbf{.816} & \underline{27.57}/\underline{.818} & \textbf{27.44}/\textbf{.819} \\
		\bottomrule
	\end{tabular}
\end{table*}

\begin{table}[t]
	\centering\scriptsize
	\caption{FMD denoising results. PSNR (dB) / SSIM are averaged over three seeds on the held-out field. Bold and underline mark the best and second-best results per metric. $^\ast$Without fine-tuning.}
	\label{tab:fmd}
	\setlength{\tabcolsep}{2.5pt}
	\begin{tabular}{lcccc}
		\toprule
		method & CF-FISH & CF-MICE & TP-MICE & mean \\
		\midrule
		BM3D~\cite{dabov2007bm3d}$^\ast$ & 27.01/.826 & 33.18/.942 & 31.63/.886 & 30.61/.885 \\
		VST\,+\,BM3D~\cite{anscombe1948,makitalo2011}$^\ast$ & 25.73/.809 & 33.62/.946 & 31.55/.889 & 30.30/.881 \\
		Noise2VST~\cite{herbreteau2025noise2vst}$^\ast$ & 28.58/.879 & 34.02/.949 & 32.14/.895 & 31.58/.908 \\
		\pbridge, $\pmo{=}0$$^\ast$ & 24.68/.811 & 31.18/.932 & 30.45/.887 & 28.77/.877 \\
		NAFNet~\cite{chen2022nafnet} & 28.23/.878 & 34.06/\underline{.951} & \underline{32.61}/\underline{.898} & 31.63/.909 \\
		Restormer~\cite{zamir2022restormer} & 28.00/.873 & \underline{34.19}/\underline{.951} & 32.39/\underline{.898} & 31.52/.908 \\
		EAMamba~\cite{lin2025eamamba} & 28.00/.874 & 34.13/\underline{.951} & 32.51/\underline{.898} & 31.55/.908 \\
		DDRM~\cite{kawar2022ddrm}$^\dagger$ & 22.67/.539 & 24.98/.546 & 24.88/.399 & 24.18/.495 \\
		I$^2$SB~\cite{liu2023i2sb} & 28.25/.867 & 34.03/.948 & 31.19/.880 & 31.15/.898 \\
		ResShift~\cite{yue2025resshift} & 27.98/.871 & 33.28/.946 & 31.70/.893 & 30.99/.903 \\
		FBI-Denoiser~\cite{byun2021fbi} & 21.67/.624 & 32.10/.888 & 30.57/.844 & 28.12/.785 \\
		VIRNet~\cite{yue2024virnet} & \textbf{28.70}/\underline{.883} & \textbf{34.24}/\textbf{.952} & 32.55/\textbf{.899} & \textbf{31.83}/\underline{.911} \\
		\pbridge, $\pmo{=}1.5$$^\ast$ & 27.47/.868 & 33.47/.944 & 31.70/.891 & 30.88/.901 \\
		\pbridge, conditional & \underline{28.59}/\textbf{.885} & 34.01/\textbf{.952} & \textbf{32.68}/\textbf{.899} & \underline{31.76}/\textbf{.912} \\
		\bottomrule
	\end{tabular}
	\par\smallskip
	{\raggedright $^\dagger$DDRM uses the realised per-image noise level.\par}
\end{table}


\begin{table}[t]
	\centering\scriptsize
	\caption{Real SAR despeckling results. $\mathcal{M}/\mathrm{RIS}$ are averaged over sensors. Lower is better. Bold and underline mark the best and second-best results per metric.}
	\label{tab:sar}
	\setlength{\tabcolsep}{2.4pt}
	\begin{tabular}{lcccc}
		\toprule
		\multirow{2}{*}{method} & \multicolumn{4}{c}{$\mathcal{M}$\,/\,$\mathrm{RIS}$, lower better} \\
		\cmidrule(lr){2-5}
		& miniSAR & FARAD-X & TerraSAR-X & mean \\
		\midrule
		BM3D~\cite{dabov2007bm3d} & 32.5/55.1 & 41.0/78.1 & 49.2/109.5 & 40.9/80.9 \\
		VST\,+\,BM3D~\cite{anscombe1948,makitalo2011} & 34.7/46.4 & 35.8/64.2 & 45.4/77.9 & 38.6/62.8 \\
		Noise2VST~\cite{herbreteau2025noise2vst} & 25.5/37.9 & 30.5/50.3 & 39.6/77.5 & 31.9/55.2 \\
		\pbridge, $\pmo{=}0$ & 36.3/63.8 & 39.9/73.6 & 46.7/100.9 & 40.9/79.4 \\
		NAFNet~\cite{chen2022nafnet} & \underline{12.0}/\underline{17.4} & \textbf{21.8}/\textbf{34.7} & \underline{32.2}/60.8 & \underline{22.0}/\underline{37.7} \\
		Restormer~\cite{zamir2022restormer} & 29.9/48.2 & 41.3/78.0 & 47.4/103.3 & 39.5/76.5 \\
		EAMamba~\cite{lin2025eamamba} & 25.4/39.6 & 31.5/53.6 & 43.4/88.9 & 33.4/60.7 \\
		DDRM~\cite{kawar2022ddrm} & 20.9/30.8 & 25.2/38.4 & 34.2/\underline{58.8} & 26.7/42.7 \\
		I$^2$SB~\cite{liu2023i2sb} & 12.7/18.5 & 30.0/50.1 & 34.3/65.6 & 25.7/44.7 \\
		ResShift~\cite{yue2025resshift} & 23.8/37.1 & 35.4/62.6 & 42.7/88.1 & 33.9/62.6 \\
		FBI-Denoiser~\cite{byun2021fbi} & 37.2/65.0 & 43.0/81.4 & 45.5/93.9 & 41.9/80.1 \\
		VIRNet~\cite{yue2024virnet} & 18.2/28.0 & 29.8/50.3 & 38.6/75.5 & 28.8/51.3 \\
		\pbridge, $\pmo{=}1.5$ & 33.3/56.4 & 43.2/83.8 & 49.6/111.7 & 42.0/84.0 \\
		\pbridge, conditional & \textbf{7.3}/\textbf{14.2} & \underline{22.4}/\underline{37.4} & \textbf{28.6}/\textbf{54.4} & \textbf{19.4}/\textbf{35.3} \\
		\bottomrule
	\end{tabular}
\end{table}

\subsection{Training objective}
\label{sec:objective}

We optimize $\mathcal L=\lambda_{\mathrm{rec}}\mathcal L_{\mathrm{rec}}+\lambda_{\mathrm{mom}}\mathcal L_{\mathrm{mom}}+\lambda_{\mathrm{cons}}\mathcal L_{\mathrm{cons}}$. The reconstruction term uses a constant schedule weight,
\begin{equation}
\begin{aligned}
  \mathcal L_{\mathrm{rec}}
  &=\E\!\left[w\,d_p(x_{\star},\hat x_{\star}(x_t;t,\xobs))\right],\\
  w&=\E_t[\nu(t)/\nuobs],
\end{aligned}
  \label{eq:l-rec}
\end{equation}
where $t$ follows the training sampling distribution. The likelihood-ratio scale gives a time-dependent weight proportional to $\nu(t)$. We replace the normalized weight $\nu(t)/\nuobs$ with its mean over training steps. This preserves the average weight and the pointwise reconstruction optimum while distributing weight uniformly across steps. The corresponding ablation is reported in \cref{tab:ablation}. The conditional model additionally uses the exponent-dependent loss weight $\muref^{\,2-p}$.

The moment term is
\begin{equation}
  \mathcal L_{\mathrm{mom}}
  =(\bar\varepsilon)^2+
    \big(\widehat{\Var}(\varepsilon)-1\big)^2,
  \label{eq:l-mom}
\end{equation}
using the Pearson residual
\begin{equation}
  \varepsilon=
  \frac{\xobs-\hat x_{\star}}{\sqrt{V(\hat x_{\star})/\nuobs}}.
  \label{eq:pearson}
\end{equation}
Under oracle predictions, each residual has zero mean and unit variance. Higher-order moments depend on the noise model. At $p=1$, residual skewness is $(\nuobs x_{\star})^{-1/2}$. At $p=2$, the term reduces to a constraint on the speckle ratio variance, whereas at $p=1$, it standardizes count residuals by their predicted Poisson variance. Spatial dependence affects the reliability of pooled estimates. For learned predictions, this term regularizes residual statistics and can shift the reconstruction optimum.

The trajectory term supervises a network evaluation at a state generated by one deterministic update:
\begin{equation}
\begin{aligned}
  \mathcal L_{\mathrm{cons}}
  &=\E\!\left[\nu(t')\,d_p(x_{\star},\hat x_{\star}(x_{t'};t',\xobs))\right],\\
  x_{t'}&=\tilde\alpha x_t+
    (1-\tilde\alpha)\operatorname{sg}(\hat x_{\star}),\\
  \tilde\alpha&=\nu(t)/\nu(t'),\qquad t'<t .
\end{aligned}
  \label{eq:l-cons}
\end{equation}
The stop-gradient operator blocks differentiation through the prediction that generates $x_{t'}$. Reconstruction optimality on the forward training distribution need not imply stationarity on this induced distribution, even when their supports overlap (\cref{sec:sup-cons}). Training settings and implementation details are provided in \cref{sec:protocol,sec:sup-impl}.

\section{Experiments}
\label{sec:experiments}

We evaluate \pbridge{} on synthetic noise spanning the exponent and severity axes, followed by real fluorescence microscopy and SAR data. The experiments examine exponent mismatch and conditioning, compare restoration methods, and evaluate noise parameter estimation and robustness to exponent.

\subsection{Experimental setup}
\label{sec:protocol}

\noindent\textbf{Training data.} We use a pool of $1{,}100$ clean natural images from BSDS500 and DIV2K~\cite{arbelaez2011bsds,agustsson2017div2k}, with $256\times256$ training crops. Clean intensities are mapped to $\XW=[0.05,1]$ using the intensity offset. Noisy observations are generated from the corresponding Tweedie model.

\noindent\textbf{Synthetic evaluation.} We evaluate on $64$ held-out crops with noise exponents $\pd\in\{1,1.25,1.5,1.75,2\}$ and severities $\eta\in\{0.25,0.5,1\}$ (mild, moderate, and severe). Each fixed-exponent model is tested across all five exponents, with results averaged over three corruption seeds. Comparisons with external methods use $\pd\in\{1,1.5,2\}$.

\noindent\textbf{Real-data evaluation.} Noise calibration covers $12$ FMD fluorescence microscopy configurations~\cite{zhang2019fmd} and three SAR sensors: miniSAR, FARAD-X, and TerraSAR-X. Restoration comparisons use three FMD configurations, following the prior protocol, and all three SAR sensors. For each FMD configuration, we use $19$ fields for fine-tuning over $20{,}000$ steps and one held-out field for evaluation. Dataset-specific preprocessing and calibration are described in the corresponding experimental subsections.

\noindent\textbf{Evaluation metrics.} We report PSNR and SSIM against clean synthetic images or the FMD frame-average reference. The supplementary sampling-budget study also reports dark-region log-RMSE (\cref{tab:s-nfe}). For real SAR, where clean references are unavailable, we report the ratio-image measures $\mathcal{M}$~\cite{gomez2017unassisted} and $\mathrm{RIS}$~\cite{vitale2019guided} alongside visual comparisons.

\noindent\textbf{Model and training.} We use a UNet with time embedding, containing $15.19$M parameters for fixed-exponent models and $15.28$M with exponent conditioning. The conditional model samples $p$ uniformly from $\{1,1.05,\ldots,2\}$ per micro-batch. Each bridge model is trained once for $50{,}000$ steps with global batch size $24$, $1{,}000$ warmup steps, and cosine learning-rate decay from $10^{-4}$ to $10^{-7}$. Loss weights are fixed at $(\lambda_{\mathrm{rec}},\lambda_{\mathrm{mom}},\lambda_{\mathrm{cons}})=(10,1,5)$. Implementation details are provided in the supplement.

\begin{figure}[t]
	\centering
	\includegraphics[width=\columnwidth]{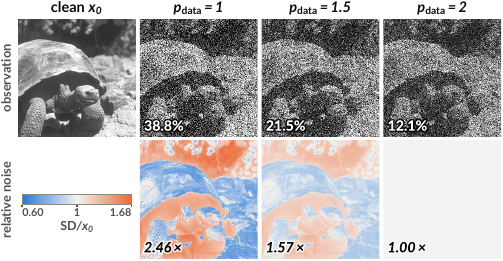}
	\caption{Effect of the noise exponent at matched relative noise. Top: clean image $x_0$ and noisy observations; percentages give the fraction of pixels at the intensity floor. Bottom: relative noise $\mathrm{SD}/x_0$, normalized at $\muref$; values give the maximum-to-minimum ratio.}
	\label{fig:corruption}
\end{figure}

\noindent\textbf{Noise calibration.} We compare different noise models by matching their relative noise at a reference intensity. We use $\muref=0.464864$, the mean intensity of the offset training pool, and define the noise level $\eta$ as the ratio of noise standard deviation to this reference intensity. The required measurement index is then
\begin{equation}
  \nuobs(p)=\eta^{-2}\muref^{\,p-2}.
  \label{eq:calib}
\end{equation}
This calibration matches noise levels at $\muref$. Away from this reference, relative noise increases towards darker regions for $p<2$, whereas it remains constant across intensities for $p=2$. \Cref{fig:corruption} illustrates how the noise exponent affects the spatial distribution of relative noise.

\noindent\textbf{Inference and severity transfer.} Unless otherwise stated, \pbridge{} uses deterministic sampling with five network evaluations. Models trained at severe noise ($\eta=1$) are also evaluated at moderate and mild noise without retraining. For a less noisy input, sampling starts at its corresponding measurement index within the trained range. For real-image evaluation, calibrated noise-plane coordinates $(p,\nu)$ configure the bridge and select the starting point for denoising.

%

\subsection{Behaviour along the exponent axis}
\label{sec:exp-axis}

\begin{figure*}[t]
	\centering
	\includegraphics[width=0.75\textwidth]{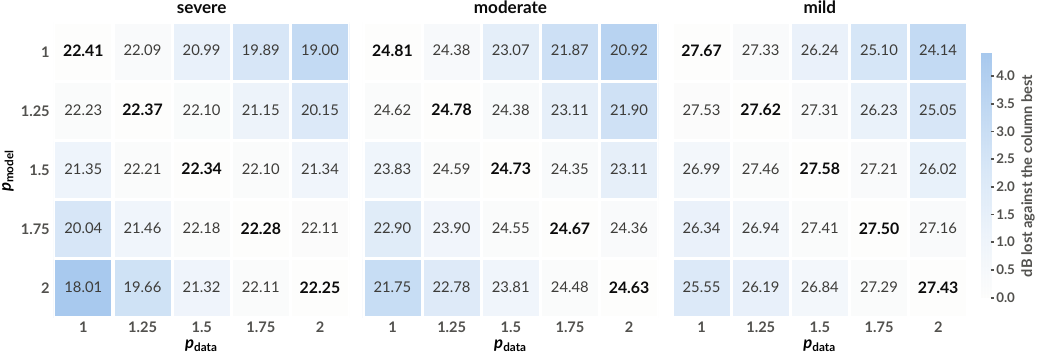}
	\caption{PSNR (dB) across model exponents (rows) and data exponents (columns), averaged over three corruption seeds at $\NFE=5$. Bold marks matched exponents, and shading shows the PSNR loss relative to the best result in each column.}
	\label{fig:mismatch}
\end{figure*}

\phantomsection\label{sec:exp-mismatch}
\noindent\textbf{The cost of a mismatched exponent.} \Cref{fig:mismatch} compares models trained at different noise exponents on the same noisy inputs. Each column fixes $\pd$, allowing direct comparison with the matched model. At severe noise, applying a Gamma-trained model ($\pmo=2$) to Poisson data ($\pd=1$) reduces PSNR by $4.40\dB$, the largest loss across the three noise levels. The reverse mismatch costs approximately $3.3\dB$, showing that the two directions are not equally harmful. Across the mismatched pairs in the $p\in\{1,1.5,2\}$ subgrid, the average loss at severe noise is $2.00\dB$.

On the $5\times5$ evaluation grid, the matched model achieves the highest PSNR in every column at all three noise levels. Performance differences remain observable even between adjacent exponents separated by $0.25$, indicating sensitivity to relatively small mismatches. Since no single fixed-exponent model is optimal across the full range, we next examine which exponent provides the best worst-case performance when the noise exponent is unknown.

\begin{figure*}[t]
	\centering
	\includegraphics[width=0.75\textwidth]{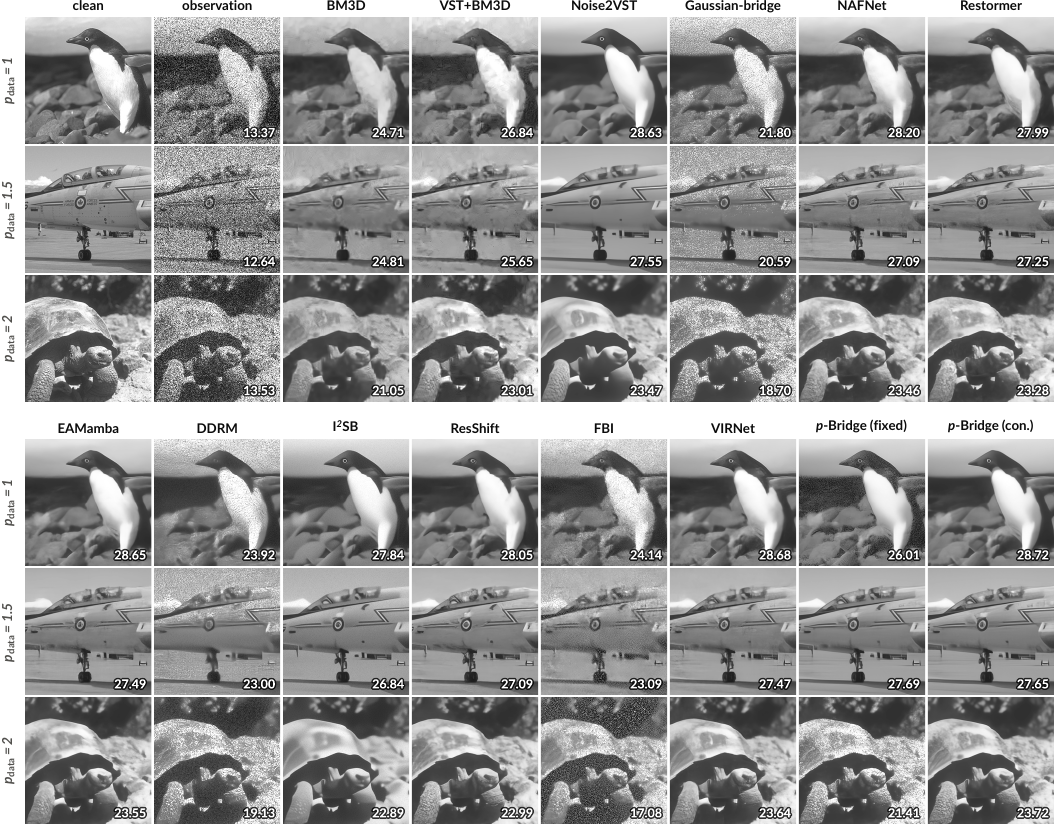}
	\caption{Synthetic-image restorations at moderate noise ($\eta=0.5$), with one row per noise exponent. Corner values show PSNR (dB).}
	\label{fig:grid-moderate}
\end{figure*}

\begin{figure*}[p]
	\centering
	\includegraphics[width=0.72\textwidth]{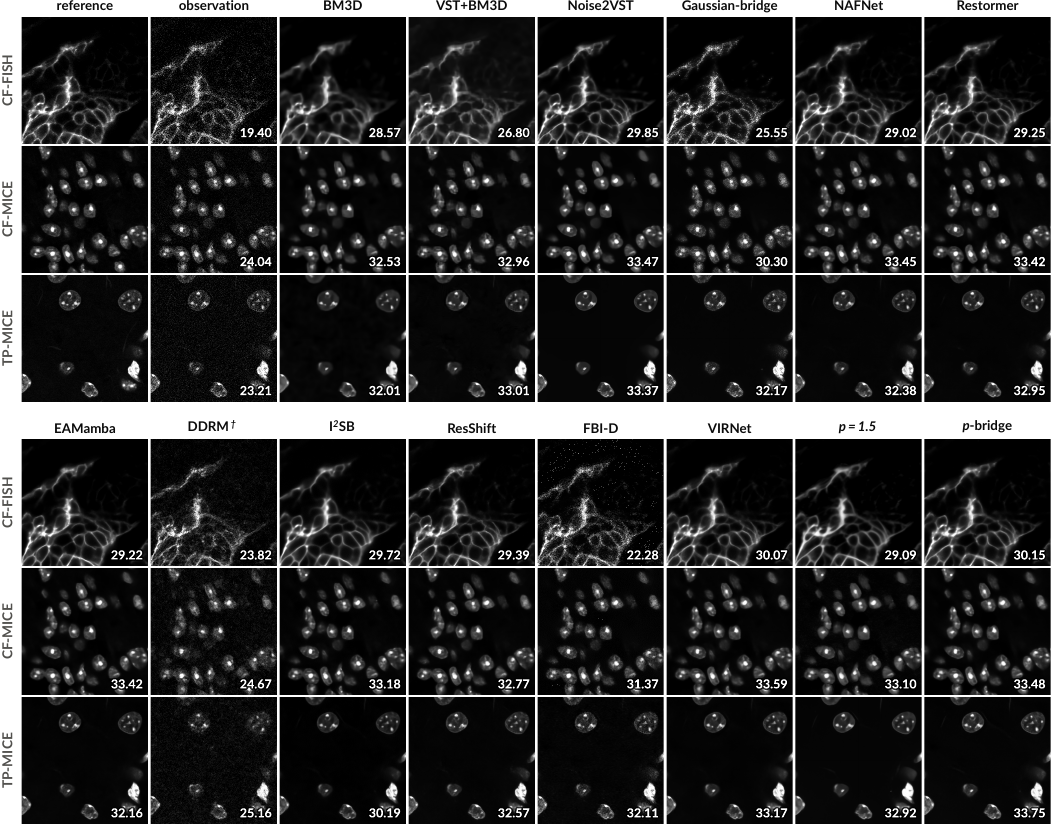}
	\caption{FMD restorations, one crop per configuration. Corner values show PSNR (dB).}
	\label{fig:fmd}
	\vspace{0.5em}
	\includegraphics[width=0.72\textwidth]{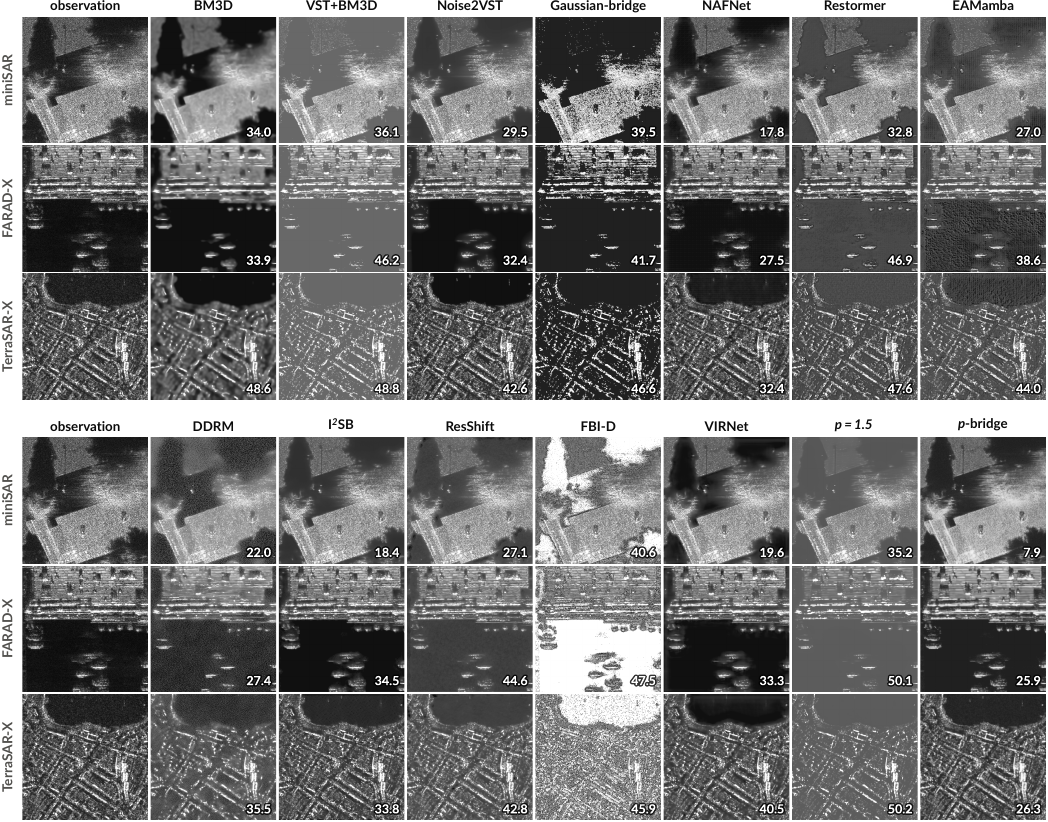}
	\caption{Real SAR restorations, one window per sensor. Corner values show $\mathcal{M}$ (lower is better).}
	\label{fig:sar}
\end{figure*}

\phantomsection\label{sec:exp-interior}
\noindent\textbf{The minimax exponent.} When the noise exponent is unknown, we evaluate each fixed-exponent model by its lowest PSNR across the tested noise exponents. \Cref{tab:cond} shows that $p=1.5$ maximizes this worst-case PSNR at severe and moderate noise, reaching $21.339$ and $23.112\dB$, respectively. At mild noise, the optimum shifts to $p=1.75$, with $26.336\dB$. Thus, an interior exponent offers the strongest worst-case performance among the tested fixed models, but no single exponent is optimal at every noise level. We therefore use $p=1.5$ as a fixed-model baseline and next examine whether conditioning on the noise exponent improves robustness across the full range.

\phantomsection\label{sec:exp-cond}
\noindent\textbf{The $p$-conditional network.} To cover different noise exponents with a single model, we condition the network on $p$ through an embedding alongside the time embedding. During training, $p$ is sampled uniformly from $\{1,1.05,\ldots,2\}$ for each micro-batch.

\Cref{tab:cond} compares the conditional network with fixed-exponent models in terms of worst-case PSNR across noise exponents. Using estimated rather than true exponents, the conditional network improves over the fixed $p=1.5$ model by $0.921$, $1.525$, and $1.413\dB$ at severe, moderate, and mild noise, respectively. Against the best fixed-exponent model at each severity, the gains are $0.921$, $1.525$, and $1.099\dB$. The smaller mild-noise gain reflects the stronger $p=1.75$ baseline.

We also compare the conditional network with an oracle-selected set of five specialist models, one for each tested exponent. At severe, moderate, and mild noise, the specialists achieve minimum PSNRs across the tested exponents of $22.254$, $24.631$, and $27.431\dB$, compared with $22.260$, $24.637$, and $27.435\dB$ for the conditional network using estimated exponents. The minimum PSNRs differ by at most $0.006\dB$ across the three noise levels.

Using estimated exponents yields nearly identical minimum PSNR to using the true exponents. In \cref{tab:cond}, the differences between these two inputs to the same conditional network are $0.000$, $+0.002$, and $-0.001\dB$ at severe, moderate, and mild noise, respectively.

To verify that the network uses the exponent input, we vary the conditioning exponent while keeping the data exponent fixed. The matching exponent achieves the highest PSNR for all five tested data exponents at every noise level, confirming that the conditioning input affects restoration performance.

\subsection{Comparison}
\label{sec:comparison}

We compare \pbridge{} with baseline methods on synthetic images, FMD fluorescence microscopy, and real SAR data. We report PSNR and SSIM for synthetic and FMD images, and ratio-image metrics with visual comparisons for SAR.

\phantomsection\label{sec:exp-synthetic}
\noindent\textbf{Synthetic images.} \Cref{tab:baselines} reports results for nine combinations of noise exponent and severity. Compared with external baselines, conditional \pbridge{} achieves the highest PSNR in eight settings and the highest SSIM in all nine. Noise2VST leads only in PSNR at mild noise with $\pd=1$, by $0.02\dB$. Improvements over VIRNet are small but consistent, ranging from $0.06$ to $0.09\dB$. Performance also varies less across noise exponents. At severe noise, PSNR drops by $0.13\dB$ from $\pd=1$ to $2$, compared with $0.94\dB$ for Noise2VST. \Cref{fig:grid-moderate} provides visual comparisons.

\phantomsection\label{sec:exp-fmd}
\noindent\textbf{Fluorescence microscopy (FMD).} \Cref{tab:fmd} reports results on three FMD configurations. Conditional \pbridge{} achieves the highest mean SSIM ($0.912$) and the second-highest mean PSNR ($31.76\dB$), compared with $31.83\dB$ for VIRNet. The PSNR ranking varies by configuration. Our model trails VIRNet by $0.11$ and $0.23\dB$ on CF-FISH and CF-MICE, respectively, but leads on TP-MICE with $32.68\dB$, a $0.13\dB$ gain. For SSIM, it ranks first on CF-FISH and shares first place with VIRNet on the other two configurations. Its mean PSNR also exceeds I$^2$SB by $0.61\dB$. Noise parameters are calibrated from training-field frame stacks. All comparisons follow the same protocol. The table distinguishes fine-tuned methods from those transferred without fine-tuning. \Cref{fig:fmd} shows the restoration results.

\phantomsection\label{sec:exp-sar}
\noindent\textbf{SAR images.} \Cref{tab:sar} reports results on miniSAR, FARAD-X, and TerraSAR-X, using eight windows per sensor. Among the ranked methods, \pbridge{} achieves the lowest mean $\mathcal{M}$ and $\mathrm{RIS}$ ($19.4/35.3$), followed by NAFNet ($22.0/37.7$). It ranks first on both metrics for miniSAR and TerraSAR-X, second for FARAD-X, where NAFNet scores lower by $0.6$ and $2.7$, respectively. \Cref{fig:sar} complements these ratio-image metrics with visual comparisons of noise removal and detail preservation. Noise calibration is detailed in the supplement.

\subsection{Ablation studies}
\label{sec:exp-component-ablation}

\begin{table}[t]
  \centering
  \fontsize{6.5}{7.5}\selectfont
  \begin{minipage}{\linewidth}
  \centering
  \caption{Component ablations at severe noise and $\NFE=5$, using the true exponent. Each row changes one setting. Minimum PSNR (dB) is taken over the tested data exponents. $\Delta$ is the difference from the full model.}
  \label{tab:ablation}
  \setlength{\tabcolsep}{4pt}
  \begin{tabular*}{\linewidth}{@{\extracolsep{\fill}}lllcc@{}}
    \toprule
    component & full model & ablation & min.\ PSNR & $\Delta$ \\
    \midrule
    Full conditional model &  &  & $22.260$ & -- \\
    \midrule
    Reconstruction weight & $\E_t[\nu(t)/\nuobs]$ & $\nu(t)/\nuobs$ & $21.916$ & $-0.345$ \\
    Exponent-dependent weight & $\muref^{\,2-p}$ & $1$ & $22.248$ & $-0.012$ \\
    Moment regularization & $\lambda_{\mathrm{mom}}=1$ & $\lambda_{\mathrm{mom}}=0$ & $22.251$ & $-0.010$ \\
    Consistency regularization & $\lambda_{\mathrm{cons}}=5$ & $\lambda_{\mathrm{cons}}=0$ & $22.051$ & $-0.210$ \\
    Index schedule & Geometric & Linear & $21.992$ & $-0.268$ \\
    \midrule
    Reverse update & $\alpha_t x_t+(1-\alpha_t)\hat x_{\star}$ & $\max(\hat x_{\star},0)$ & $22.075$ & $-0.186$ \\
    Output-range projection & Enabled & Disabled & $22.260$ & $-0.000$ \\
    \bottomrule
  \end{tabular*}
  \par\smallskip
  \begin{minipage}{\linewidth}
    \footnotesize Reconstruction weights have the same mean over sampled training steps, and loss coefficients are unchanged in this comparison. In the reverse update, $\alpha_t=\nu(t)/\nu(t-1)$. The ablated update replaces the state with the current prediction, with nonnegative clipping.
  \end{minipage}
  \end{minipage}
\end{table}

\begin{figure}[t]
	\centering
	\includegraphics[width=\columnwidth]{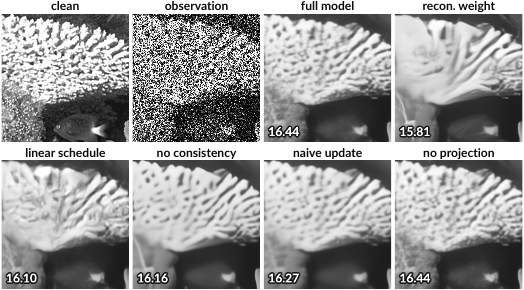}
	\caption{Qualitative component ablation at severe noise, $\pd=1.5$. Corner values report PSNR (dB).}
	\label{fig:ablate-qual}
\end{figure}

\begin{figure}[t]
	\centering
	\includegraphics[width=\columnwidth]{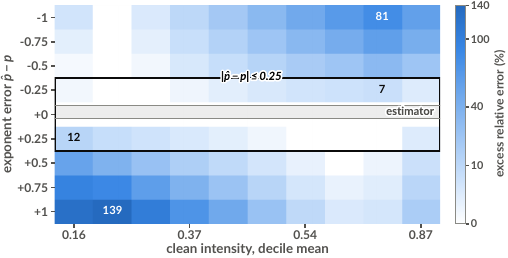}
	\caption{Excess relative error over the matched-exponent model, by intensity decile and signed exponent error. The box marks $|\hat p-p|\leq0.25$, and the gray strip shows the range of the estimator errors. Rows average exponent pairs with the same error.}
	\label{fig:estimator-mismatch}
\end{figure}

\begin{table}[t]
	\centering
	\scriptsize
	\caption{PSNR gains (dB) over the fixed $p=1.5$ model under worst-case exponent errors. The conditioning exponent satisfies $|\hat p-\pd|\leq\delta$ within the tested exponent range. $\delta=0$ uses the true exponent, while $\delta=\infty$ removes the error constraint. Positive values favor the conditional model.}
	\label{tab:s-delta}
	\setlength{\tabcolsep}{4pt}
	\begin{tabular}{lcccc}
		\toprule
		vs.\ fixed $p{=}1.5$ & $\delta{=}0$ & $\delta{=}0.25$ & $\delta{=}0.5$ & $\delta{=}\infty$ \\
		\midrule
		severe   & $+0.922$ & $+0.830$ & $+0.107$ & $-1.707$ \\
		moderate & $+1.523$ & $+1.311$ & $+0.184$ & $-1.863$ \\
		mild     & $+1.414$ & $+1.233$ & $+0.177$ & $-1.796$ \\
		\bottomrule
	\end{tabular}
\end{table}

\noindent\textbf{Evaluation protocol.} \Cref{tab:ablation} summarizes the component ablations of the conditional model. Each variant changes one setting while keeping the others fixed. Training-side variants are retrained with the same data and budget, whereas inference-only variants reuse the full model's checkpoint.

\noindent\textbf{Loss weighting and index schedule.} Measurement-dependent weighting and a linear index schedule reduce PSNR by $0.345$ and $0.268\dB$, respectively. Although the mean reconstruction weight is unchanged, measurement-dependent weighting assigns greater weight to cleaner states. Likewise, a linear schedule allocates fewer training steps near the noisy observation than a geometric schedule over the same index range. These results support the use of constant weighting and geometric spacing to emphasize noisy states under a fixed training budget.

\noindent\textbf{Regularization and reverse update.} At five network evaluations, removing consistency regularization and replacing the reverse mixture with direct prediction reduce PSNR by $0.210$ and $0.186\dB$, respectively. These losses increase to $0.452$ and $1.654\dB$ at $25$ evaluations. At one evaluation, no reverse update is applied, so the two samplers produce identical outputs. Removing consistency regularization changes PSNR by only $0.004\dB$. By comparison, removing moment regularization or exponent-dependent loss weighting reduces PSNR by only $0.010$ and $0.012\dB$, respectively.

\Cref{fig:ablate-qual} illustrates the structural degradation of ablations. No-projection result remains visually close to the full model.

\noindent\textbf{Output-range projection.} With matched exponents, removing projection changes PSNR by at most $0.001\dB$ at one, five, and $25$ network evaluations. Its benefit is more evident when the conditioning exponent is below the data exponent. For Gamma noise ($\pd=2$) with the conditioning exponent set to $1$, removing projection reduces PSNR by $0.223$, $0.168$, and $0.139\dB$, respectively. Projection therefore provides a modest benefit under exponent mismatch, with little effect on matched-noise performance.

\subsection{Noise parameter estimation and robustness}
\label{sec:exp-ablation}

\begin{figure}[t]
	\centering
	\includegraphics[width=0.9\columnwidth]{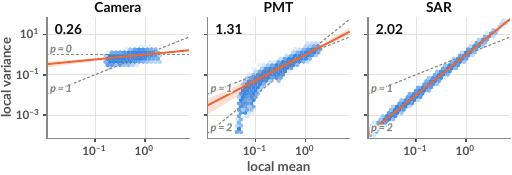}
	\caption{Variance--mean relations by detector class. Each configuration is normalized by its fitted variance at the median signal. Dashed lines mark $p=0,1,2$; shaded bands span the fitted exponents, and corner values give the class mean.}
	\label{fig:ptc}
\end{figure}

\begin{figure}[t]
	\centering
	\includegraphics[width=0.9\columnwidth]{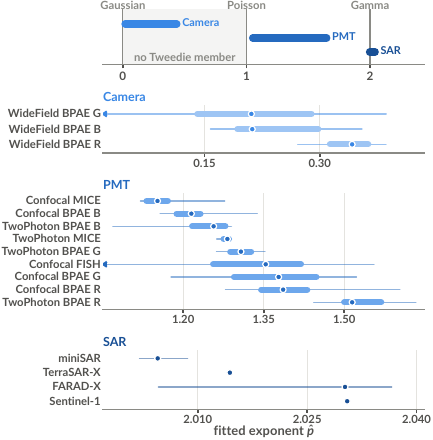}
	\caption{Fitted exponents by imaging configuration and detector class. Dots show medians, thick bars interquartile ranges, and thin lines full ranges; carets indicate values outside the displayed range.}
	\label{fig:pdist}
\end{figure}

\begin{figure}[t]
	\centering
	\includegraphics[width=0.9\columnwidth]{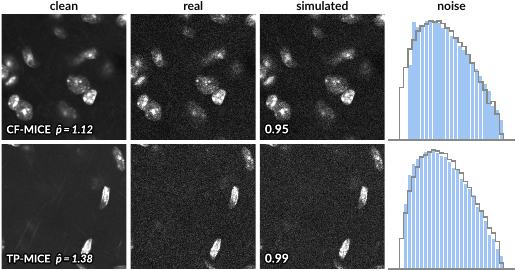}
	\caption{Real and simulated FMD frames using the fitted exponent and index. Rows show the clean signal, real and simulated observations, and middle-band noise histograms (real filled; simulated outlined). Numbers are simulated-to-real noise-standard-deviation ratios averaged over the dark and bright bands; values closer to $1$ indicate better agreement.}
	\label{fig:simreal}
\end{figure}

\begin{figure}[t]
	\centering
	\includegraphics[width=0.9\columnwidth]{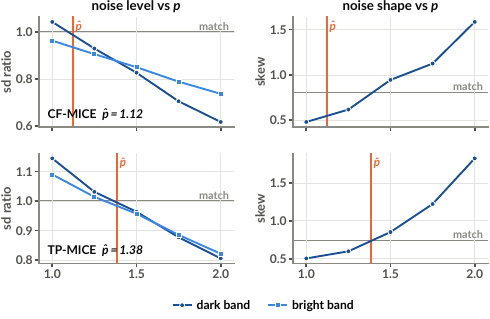}
	\caption{Noise statistics versus candidate exponent for the fields in \cref{fig:simreal}. Left: simulated-to-real noise-standard-deviation ratios; right: simulated skewness, for dark and bright bands. Curves closer to the horizontal match lines agree better with real data; vertical lines mark the fitted exponents.}
	\label{fig:psweep}
\end{figure}

\phantomsection\label{sec:exp-delta}
\noindent\textbf{Tolerance to exponent estimation error.} We first examine how accurately the exponent must be estimated to retain the benefit of conditioning. \Cref{tab:s-delta} compares the conditional network with the fixed $p=1.5$ model under worst-case errors satisfying $|\hat p-\pd|\leq\delta$. At $\delta=0.25$, the PSNR gains remain $0.830$--$1.311\dB$ across the three noise levels, only $0.09$--$0.21\dB$ below exact conditioning. At $\delta=0.5$, the gains decrease to $0.107$--$0.184\dB$ but remain positive. Without an error constraint over the tested exponent range, the conditional model falls below the fixed baseline by $1.707$--$1.863\dB$. Thus, errors within $0.25$ preserve most of the conditioning benefit in these tests.

\Cref{fig:estimator-mismatch} shows how this cost varies with intensity. Overestimating the exponent concentrates excess error in dark regions, whereas underestimating it affects bright regions. The observed estimator errors remain within $\delta=0.25$.

\phantomsection\label{sec:exp-phat}
\noindent\textbf{Noise parameter estimation.} We estimate the noise parameters directly from the noisy image without additional training. Taking logarithms of $\Var(\xobs\mid\mu)=\mu^p/\nu$ gives a linear relation with slope $p$ and intercept $-\log\nu$. We fit a robust regression of estimated log-noise variance on log-mean intensity. The fitted slope and intercept jointly determine $p$ and $\nu$, with dispersion absorbed into the measurement index. Details of local-statistic estimation and regression are provided in \cref{sec:obj-phat}.

On the evaluation grid of five exponents, three noise levels, and three corruption seeds, the reported worst-case absolute error in $p$ is $0.0914$, below the $0.25$ target identified above. The median relative error in $\nu$ ranges from $2.8\%$ to $4.1\%$. Mild noise is more challenging to estimate because, at a fixed exponent, $\nu$ is $16$ times its severe-noise value, leaving a weaker noise signal. Additional tests of estimation bias and sensitivity to estimator settings are provided in \cref{sec:sup-estimator-robustness}.

\noindent\textbf{Estimates outside the supported range.} The regression does not constrain $\hat p$ to $[1,2]$. In five of the $45$ evaluated configurations, estimates fall outside this interval by at most $0.032$, all at $\pd=1$ or $2$. We clip $\hat p$ to $[1,2]$ to match the range supported by the conditional model and its bridge implementation. The blind results in \cref{tab:cond} use these clipped estimates.

\noindent\textbf{Real-data calibration.} \Cref{fig:ptc} shows the variance--mean relation used to fit the real-data exponents, and \cref{fig:pdist} summarizes the estimates across imaging configurations. The fitted slopes differ across detector classes and configurations, supporting configuration-specific calibration rather than a single fixed exponent.

\noindent\textbf{Calibration check on FMD.} To check whether the fitted law reproduces the observed noise, \cref{fig:simreal} compares real and simulated FMD frames at matched clean signal, exponent, and index. \Cref{fig:psweep} then varies the candidate exponent while refitting the corresponding index, showing how the simulated noise level and skew change around the fitted value.

\section{Conclusion}

\pbridge{} connects statistical noise modeling with the design of a denoising trajectory. Measurement accumulation provides the common bridge structure, while exponent conditioning adapts the learned prediction to the noise law. The experiments show that this adaptation matters beyond selecting a noise level. Fixed-exponent models lose accuracy under mismatched noise laws, whereas a shared conditional network achieves cross-exponent robustness comparable to oracle-selected specialists. Using estimated exponents leaves this robustness nearly unchanged, and moderate conditioning errors preserve most of the benefit. Results on synthetic images, fluorescence microscopy, and SAR further support the usefulness of the formulation across imaging regimes. Together, these findings support adapting to different Tweedie noise models by making the noise law an explicit condition within a shared restoration model, without a separate checkpoint for each exponent.


\bibliographystyle{IEEEtran}
\bibliography{main}


\clearpage
\setcounter{page}{1}
\maketitlesupplementary

%
%
%
\setcounter{section}{0}
\setcounter{table}{0}
\setcounter{figure}{0}
\setcounter{equation}{0}
\renewcommand{\thesection}{S\arabic{section}}
\renewcommand{\thetable}{S\arabic{table}}
\renewcommand{\thefigure}{S\arabic{figure}}
\renewcommand{\theequation}{S\arabic{equation}}
\setcounter{algorithm}{0}
\renewcommand{\thealgorithm}{S\arabic{algorithm}}

\makeatletter
\let\supp@addcontentsline\addcontentsline
\renewcommand{\addcontentsline}[3]{%
  \def\supp@extension{#1}%
  \def\supp@toc{toc}%
  \ifx\supp@extension\supp@toc
    \supp@addcontentsline{stoc}{#2}{#3}%
  \else
    \supp@addcontentsline{#1}{#2}{#3}%
  \fi
}
\newcommand{\supplementcontents}{%
  \section*{Contents}%
  \begingroup
    \small
    \setcounter{tocdepth}{2}%
    \renewcommand{\l@section}[2]{%
      \addpenalty{\@secpenalty}\addvspace{0.6em}%
      \begingroup\bfseries
        \@dottedtocline{1}{0em}{2.7em}{##1}{##2}%
      \endgroup\nobreak
    }%
    \renewcommand{\l@subsection}{\@dottedtocline{2}{2.7em}{3.5em}}%
    \@starttoc{stoc}%
  \endgroup
}
\makeatother


\supplementcontents
\twocolumn
\linespread{0.95}\selectfont

\section{Measurement model and assumptions}
\label{sec:sup-reg}

\subsection{Accumulated signal and intensity}

The measurement model is defined pixelwise by a clean mean and a measurement index $\nu$. We derive the moment identities in \cref{eq:ed-moments} from the additive density in \cref{eq:ed-density}. The natural parameter $\theta$ lies in the interior of its domain $\Theta$. For $s$ such that $\theta+s\in\operatorname{int}\Theta$, normalization of the density gives
\begin{align*}
  \E[e^{sZ}]
  &=\int e^{sz}c^{*}(z;\nu)
    e^{\theta z-\nu\kappa(\theta)}\,\mathrm{d}m_\nu(z)\\
  &=e^{\nu[\kappa(\theta+s)-\kappa(\theta)]}
    \int f_Z(z;\theta+s,\nu)\,\mathrm{d}m_\nu(z)\\
  &=e^{\nu[\kappa(\theta+s)-\kappa(\theta)]}.
\end{align*}
Here $m_\nu$ denotes the dominating measure. Taking the logarithm yields the cumulant-generating function
\begin{equation}
  K_Z(s)=\nu[\kappa(\theta+s)-\kappa(\theta)].
  \label{eq:s-cgf}
\end{equation}
Its first two derivatives at zero give the mean and variance in \cref{eq:ed-moments}. Using the mean map and variance function defined in \cref{eq:ed-maps}, normalizing the accumulated signal gives the intensity $x=Z/\nu$, also written in reproductive form as $x\sim\ED(\mu,1/\nu)$. Therefore,
\begin{align*}
  \E[x]&=\frac{\E[Z]}{\nu}=\mu,\\
  \Var(x)&=\frac{\Var(Z)}{\nu^2}
           =\frac{V(\mu)}{\nu}.
\end{align*}
Setting $\mu=x_{\star}$ recovers the conditional measurement moments in \cref{eq:fwd}. The power variance function $V(\mu)=\mu^p$ gives the Tweedie family~\cite{tweedie1984}. We consider $p\in\{0\}\cup[1,2]$, covering Gaussian, Poisson, compound Poisson--Gamma, and Gamma models. These distributions provide both the required moments and the closure property used to construct the bridge.

\Cref{tab:ed-members} lists the distributional functions for these models. They follow by integrating $\mathrm{d}\mu/\mathrm{d}\theta=V(\mu)$ and $\mathrm{d}\kappa/\mathrm{d}\theta=\mu$, with conventional choices of integration constants. Here $b=(1-p)\theta>0$ for $1<p<2$. The natural parameter ranges over $\mathbb{R}$ for $p=0,1$ and is negative for $1<p\leq2$.

The measurement index incorporates the noise scale. If a physical measurement amount $a$ gives variance $\phi V(\mu)/a$ for a fixed dispersion scale $\phi>0$, setting $\nu=a/\phi$ recovers $V(\mu)/\nu$. Thus, calibration determines the conversion from physical measurement amount to index.

\subsection{Conditions for the bridge construction}

The bridge uses the following assumptions.
\begin{itemize}[leftmargin=3.2em,itemsep=3pt,topsep=4pt]
  \item[(A1)] The additive family is infinitely divisible and defined for every index $\nu\in\Lambda=(0,\infty)$.
  \item[(A2)] The mean map $\tau=\kappa'$ is a bijection from $\operatorname{int}\Theta$ onto the mean domain $\Omega$. Hence
    \[
      \theta_{\star}=\tau^{-1}(x_{\star})
    \]
    exists and is unique for each $x_{\star}\in\Omega$.
  \item[(A3)] The schedule is strictly decreasing, with
    \[
      0<\nuobs=\nu(T-1)<\cdots<\nu(0)=\numax<\infty,
    \]
    so every reverse step has a positive increment
    \[
      \Delta_t=\nu(t-1)-\nu(t)>0,
      \qquad t=1,\ldots,T-1.
    \]
  \item[(A4)] Conditional on $x_{\star}$, the initial reverse state $Z_{T-1}$ and all added increments are mutually independent and share the natural parameter $\theta_{\star}$. Each earlier state is formed by adding the next increment to the accumulated state.
  \item[(A5)] All distributional identities and moment statements are interpreted pixelwise, with variances referring to individual pixel intensities.
\end{itemize}

Condition (A1) permits arbitrary positive measurement increments. Without it, a family may allow only a discrete index set. For example, a binomial model parameterized by its number of trials requires integer indices and cannot support an arbitrary real-valued measurement schedule. Condition (A4) supplies the independence needed for superposition. It does not assert independence between the bridge states, which share accumulated measurements.

\begin{table}[!t]
	\centering
	\caption{Distributional functions of the supported Tweedie models.}
	\label{tab:ed-members}
	\setlength{\tabcolsep}{2pt}
	\renewcommand{\arraystretch}{1.35}
	\footnotesize
	\begin{tabular*}{\columnwidth}{@{\extracolsep{\fill}}lcccc@{}}
		\toprule
		& Gaussian                & Poisson      & Poisson--Gamma             & Gamma             \\
		$p$              & $0$                     & $1$          & $(1,2)$                    & $2$               \\
		\midrule
		$\kappa(\theta)$ & $\sigma^{2}\theta^{2}/2$ & $e^{\theta}$ & $b^{(2-p)/(1-p)}\!/(2-p)$ & $-\log(-\theta)$  \\
		$\tau(\theta)$   & $\sigma^{2}\theta$      & $e^{\theta}$ & $b^{1/(1-p)}$              & $-1/\theta$       \\
		$V(\mu)$         & $\sigma^{2}$            & $\mu$        & $\mu^{p}$                  & $\mu^{2}$         \\
		$\tau^{-1}(\mu)$ & $\mu/\sigma^{2}$        & $\log\mu$    & $\mu^{1-p}\!/(1-p)$        & $-1/\mu$          \\
		\bottomrule
	\end{tabular*}
\end{table}

\subsection{Clean limit and intensity domains}

For a fixed clean intensity with finite $V(x_{\star})$, unbiasedness gives
\begin{align*}
  \E[(x_\nu-x_{\star})^2\mid x_{\star}]
  &=\Var(x_\nu\mid x_{\star})\\
  &=\frac{V(x_{\star})}{\nu}
    \longrightarrow 0 \qquad (\nu\to\infty).
\end{align*}
This establishes the conditional mean-square limit stated in the main paper. At finite $\numax$, the initial bridge state instead has variance $V(x_{\star})/\numax$ and is not identified with the clean image.

The mean domain, the clean-image range, and the observation support serve different roles. Under (A2), the positive Tweedie branches have $\Omega=(0,\infty)$, so their clean means are strictly positive but have no model-imposed upper bound. The implementation uses the normalized clean-image range $\XW=[\epsilon,1]$, $\epsilon>0$, and clips predicted clean images to this range. This is an implementation convention, and sampled states are not clipped. In particular, for $1<p<2$, observations can equal zero or exceed the clean-image range.

\section{Superposition and the reverse mean}
\label{sec:sup-preserve}

We prove \cref{thm:superposition,thm:reverse} and derive the oracle preservation properties.

\subsection{Closure under measurement accumulation}

Consider independent accumulated signals $Z_1\sim\EDs(\theta,\nu_1)$ and $Z_2\sim\EDs(\theta,\nu_2)$ with the same natural parameter. Independence makes their cumulant-generating functions additive. Using \cref{eq:s-cgf},
\begin{align*}
  K_{Z_1+Z_2}(s)
  &=K_{Z_1}(s)+K_{Z_2}(s)\\
  &=\nu_1[\kappa(\theta+s)-\kappa(\theta)]\\
  &\quad+\nu_2[\kappa(\theta+s)-\kappa(\theta)]\\
  &=(\nu_1+\nu_2)[\kappa(\theta+s)-\kappa(\theta)].
\end{align*}
The result is the cumulant-generating function of $\EDs(\theta,\nu_1+\nu_2)$, proving \cref{eq:add}. A shared natural parameter ensures closure. With different parameters, the sum generally does not have this form.

\subsection{The superposition coupling}

To prove \cref{thm:superposition}, condition on the clean image. By (A1), an increment with index $\Delta_t$ exists; by (A4), it shares $\theta_{\star}$ with $Z_t$ and is independent of $Z_t$. Applying \cref{eq:add} and using $\nu(t)+\Delta_t=\nu(t-1)$ gives
\[
  Z_{t-1}=Z_t+Y_{t\to t-1}\sim\EDs\big(\theta_{\star},\nu(t-1)\big),
\]
This is the marginal at $t-1$ prescribed by \cref{eq:bridge}.

The intensity-domain form follows by substituting $Z_t=\nu(t)\,x_t$ and
dividing through by $\nu(t-1)$:
\begin{align*}
  x_{t-1}
  &=\frac{Z_{t-1}}{\nu(t-1)}\\[4pt]
  &=\frac{\nu(t)\,x_t+Y_{t\to t-1}}{\nu(t-1)}\\[4pt]
  &=\frac{\nu(t)}{\nu(t-1)}\,x_t+\frac{1}{\nu(t-1)}\,Y_{t\to t-1}\\[4pt]
  &=\alpha_t\,x_t+\frac{1}{\nu(t-1)}\,Y_{t\to t-1} ,
\end{align*}
which is \cref{eq:superposition}.

\subsection{Exponent-independent mixing coefficient}

For \cref{thm:reverse}, condition on the current state and prediction $(x_t,\hat x_{\star})$. The model draws an increment with natural parameter $\hat\theta_{\star}=\tau^{-1}(\hat x_{\star})$ and index $\Delta_t$. The first moment in \cref{eq:ed-moments} gives
\[
  \E\big[Y_{t\to t-1}\mid x_t,\hat x_{\star}\big]
  =\Delta_t\,\kappa'(\hat\theta_{\star})
  =\Delta_t\,\hat x_{\star} ,
\]
so taking the conditional expectation of \cref{eq:superposition} given
$(x_t,\hat x_{\star})$,
\begin{align*}
  \E\big[x_{t-1}\mid x_t,\hat x_{\star}\big]
  &=\alpha_t\,x_t+\frac{\Delta_t}{\nu(t-1)}\,\hat x_{\star}\\[4pt]
  &=\alpha_t\,x_t+\left(1-\frac{\nu(t)}{\nu(t-1)}\right)\hat x_{\star}\\[4pt]
  &=\alpha_t\,x_t+(1-\alpha_t)\,\hat x_{\star} ,
\end{align*}
where $\Delta_t/\nu(t-1)=1-\alpha_t$. The mixing coefficient depends only on the measurement indices, whereas the increment distribution and network prediction can depend on $p$. The result gives the conditional mean of the specified plug-in transition.

\subsection{Measurement-weighted intensity averaging}

The normalized form of \cref{eq:add} gives the measurement-weighted averaging rule. If $x_i\sim\ED(\mu,1/w_i)$ are independent with positive measurement weights $w_i$, put
$Z_i:=w_ix_i$, so that $Z_i\sim\EDs(\theta,w_i)$ with the common
$\theta=\tau^{-1}(\mu)$. Then \cref{eq:add} gives
\[
  \sum_i Z_i \;\sim\; \EDs\Big(\theta,\ \sum_i w_i\Big),
\]
and dividing by $\sum_i w_i$ converts back to the reproductive form:
\begin{equation}
  \frac{\sum_i w_i x_i}{\sum_i w_i}
  \;\sim\;
  \ED\!\left(\mu,\ \frac{1}{\sum_i w_i}\right).
  \label{eq:s-avg}
\end{equation}
Thus the weighted average remains in the same family, with mean $\mu$ and measurement index $\sum_i w_i$. For measurements of a fixed scene, take $\mu=x_{\star}$ and interpret independence conditionally on that scene. Exposure, dose, or number of looks can serve as the index when the corresponding sensor model and calibration satisfy these assumptions.

\subsection{Oracle preservation properties}

We now establish the preservation properties under oracle predictions $\hat x_{\star}=x_{\star}$.

First, the stochastic update preserves the prescribed bridge marginals by \cref{thm:superposition}. The deterministic update preserves their mean. Taking the expectation of \cref{eq:reverse} over $x_t\sim\ED(x_{\star},1/\nu(t))$ gives
\[
  \begin{aligned}
  \E[x_{t-1}]
  &=\alpha_t\,\E[x_t]+(1-\alpha_t)\,x_{\star}\\
  &=\alpha_t x_{\star}+(1-\alpha_t)x_{\star}=x_{\star} .
  \end{aligned}
\]
Induction extends this mean-preservation result along the schedule. Replacing the random increment by its mean reduces the variance, so the deterministic update does not preserve the bridge distribution. For a bridge-distributed input,
\[
  \Var(x_{t-1}^{\mathrm{det}}\mid x_{\star})
  =\alpha_t^2\frac{V(x_{\star})}{\nu(t)}
  =\alpha_t\frac{V(x_{\star})}{\nu(t-1)},
\]
which is smaller than the target variance $V(x_{\star})/\nu(t-1)$.

Second, the stochastic sampler reproduces the joint trajectory of the superposition coupling. Starting from $Z_{T-1}$, each earlier state is generated by an increment independent of all states at times $\ge t$. Therefore,
\[
  q(z_{t-1}\mid z_{t:T-1},x_{\star})=q(z_{t-1}\mid z_t,x_{\star}),
\]
so the chain rule collapses to
\begin{equation}
  q(z_{T-1},\dots,z_0\mid x_{\star})
  = q(z_{T-1}\mid x_{\star})\prod_{t=1}^{T-1} q(z_{t-1}\mid z_t,x_{\star}),
  \label{eq:s-markov}
\end{equation}
Thus, the coupling is Markov in the reverse-time index, with the stochastic transition in \cref{thm:superposition}. Initialization from $Z_{T-1}\sim\EDs(\theta_{\star},\nuobs)$ recovers the full joint law. For a fixed observation $Z_{T-1}=z_{\mathrm{obs}}$, the same transitions instead generate the trajectory conditional on that observation.

These properties concern the superposition coupling. The marginals in \cref{eq:bridge} alone do not determine a unique joint law across time.

\section{Unit deviance and conditional-mean prediction}
\label{sec:sup-loss}

The Bregman representation in \cref{eq:bregman} provides a shared reconstruction objective across noise models. Let $\psi:=\kappa^{*}$. Convex conjugacy gives
\begin{equation}
  \psi(\mu)=\mu\,\tau^{-1}(\mu)-\kappa\big(\tau^{-1}(\mu)\big),
  \qquad
  \psi'(\mu)=\tau^{-1}(\mu)=\theta .
  \label{eq:s-conj}
\end{equation}

\subsection{Bregman representation}

Write $\tilde\theta=\tau^{-1}(y)$ and $\theta=\tau^{-1}(\mu)$, and expand the
Bregman divergence generated by $\psi$:
\begin{align*}
  B_\psi(y,\mu)
  &= \psi(y)-\psi(\mu)-\psi'(\mu)\,(y-\mu)\\[2pt]
  &= \big[y\tilde\theta-\kappa(\tilde\theta)\big]
    -\big[\mu\theta-\kappa(\theta)\big]
    -\theta\,(y-\mu)\\[2pt]
  &= y\tilde\theta-\kappa(\tilde\theta)-\mu\theta+\kappa(\theta)
    -\theta y+\theta\mu\\[2pt]
  &= y(\tilde\theta-\theta)-\kappa(\tilde\theta)+\kappa(\theta)\\[2pt]
  &= \tfrac12\,d(y,\mu),
\end{align*}
The second line substitutes \cref{eq:s-conj}. Cancelling the $\mu\theta$ terms then gives half the unit deviance.

\subsection{Mean-optimal reconstruction}

We first derive the pointwise gradient, then establish \cref{thm:mean} by an expectation decomposition. Using $\psi''(\mu)=1/V(\mu)$,
\begin{align*}
  \partial_\mu B_\psi(y,\mu)
  &= \partial_\mu\big[\psi(y)-\psi(\mu)-\psi'(\mu)(y-\mu)\big]\\[2pt]
  &= -\psi'(\mu)-\psi''(\mu)(y-\mu)+\psi'(\mu)\\[2pt]
  &= -\psi''(\mu)\,(y-\mu)\\[2pt]
  &= -\frac{y-\mu}{V(\mu)} ,
\end{align*}
The two $\psi'(\mu)$ terms cancel by the product rule. When differentiation and expectation can be interchanged,
\begin{equation}
  \frac{d}{d\mu}\,\E\big[d(Y,\mu)\big]
  = -\frac{2}{V(\mu)}\big(\E[Y]-\mu\big),
  \label{eq:s-meangrad}
\end{equation}
Since $V(\mu)>0$ on the mean domain, the only stationary point is $\mu=\E[Y]$. To establish optimality without interchanging differentiation and expectation, let $m:=\E[Y]$ and expand \cref{eq:bregman} about $m$:
\begin{equation}
  \E\big[d(Y,\mu)\big]=\E\big[d(Y,m)\big]+d(m,\mu),
  \qquad \mu\in\Omega ,
  \label{eq:s-bregdecomp}
\end{equation}
The cross term has zero expectation because it is linear in $Y-m$. If $m\in\Omega$ and $\E[d(Y,m)]<\infty$, strict convexity of $\psi$ makes $d(m,\mu)$ uniquely minimal at $\mu=m$. These conditions establish \cref{thm:mean}. In our experiments, $x_{\star}\in\XW=[\epsilon,1]$, so the conditional mean remains in the mean domain and the expected deviance is finite. Applying the decomposition conditionally gives
\[
  \arg\min_{\hat x_{\star}}\E\big[d(x_{\star},\hat x_{\star})\mid x_t,\xobs\big]
  =\E\big[x_{\star}\mid x_t,\xobs\big],
\]
Thus, the reconstruction objective predicts the conditional mean used by the reverse update in \cref{eq:reverse}.

\subsection{Tweedie deviance formulas}

Substituting $V(\mu)=\mu^{p}$ yields the gradient in \cref{eq:dp-grad}:
\[
  \partial_\mu d_p(y,\mu)=-\frac{2(y-\mu)}{\mu^{p}}
  =-2y\mu^{-p}+2\mu^{1-p},
\]
and integrating in $\mu$, with the constant of integration fixed by
$d_p(y,y)=0$, gives
\begin{equation}
  d_p(y,\mu)
  = 2\left[
      \frac{y^{2-p}}{(1-p)(2-p)}
      -\frac{y\,\mu^{1-p}}{1-p}
      +\frac{\mu^{2-p}}{2-p}
    \right]
  \label{eq:s-dp}
\end{equation}
for $p\notin\{0,1,2\}$, together with
\begin{equation}
  d_p(y,\mu)=
  \begin{cases}
    (y-\mu)^{2}, & p=0,\\[5pt]
    2\big[y\log(y/\mu)-(y-\mu)\big], & p=1,\\[5pt]
    2\big[y/\mu-\log(y/\mu)-1\big], & p=2,
  \end{cases}
  \label{eq:s-dp-special}
\end{equation}
The Gaussian case follows by setting $p=0$ in \cref{eq:s-dp}. The Poisson and Gamma cases follow from the limits $p\to1$ and $p\to2$, where the power terms yield logarithms.

\section{Measurement thinning kernels}
\label{sec:sup-kernels}

The forward transition reduces the measurement amount through the thinning kernel $q(z_t\mid z_{t-1};\theta_{\star})$ in \cref{eq:thin}. We derive this kernel, give its Gaussian, Poisson, and Gamma forms, and establish the latent construction in \cref{prop:latent}. We then relate the compound model to the excess-noise factor in \cref{eq:excess}.

\subsection{Independence from the clean-signal parameter}

For $p=1$, the density in \cref{eq:ed-density} is a mass function with respect to counting measure. For $1<p<2$, use the parameter-independent dominating measure $\delta_0+\mathrm{Leb}_{(0,\infty)}$. Under the canonical compound parameterization, $c^*(0;\nu)=1$, while the probability mass at zero is $\Pr(Z_\nu=0)=e^{-\nu\lambda}$. The latter is part of the probability law, not part of the dominating measure.

Conditioning on a positive total $z$ gives a mixed kernel on $[0,z]$ with a Lebesgue density on $(0,z)$ and atoms at both $0$ and $z$. Equation~\eqref{eq:thin} gives its density with respect to $\delta_0+\mathrm{Leb}_{(0,z)}+\delta_z$. In particular,
\begin{align*}
  \Pr(Z_t=0\mid Z_{t-1}=z)
  &=\frac{c^*(z;\Delta_t)}{c^*(z;\nu(t-1))},\\
  \Pr(Z_t=z\mid Z_{t-1}=z)
  &=\frac{c^*(z;\nu(t))}{c^*(z;\nu(t-1))}.
\end{align*}
For $z=0$, the conditional law is a point mass at zero. The latent construction below realizes all these components without evaluating the compound density.

By (A4) the pair $(Z_t,\,Z_{t-1}-Z_t)$ consists of independent variables at the
same $\theta_{\star}$, with indices $\nu(t)$ and $\Delta_t$. Its joint density is
therefore the product of two copies of \cref{eq:ed-density},
\[
  c^{*}\big(z_t;\nu(t)\big)\,
  c^{*}\big(z_{t-1}-z_t;\Delta_t\big)\,
  e^{\theta_{\star} z_{t-1}-\nu(t-1)\kappa(\theta_{\star})},
\]
where $z_t+(z_{t-1}-z_t)=z_{t-1}$ and $\nu(t)+\Delta_t=\nu(t-1)$. Dividing by the marginal density of $Z_{t-1}$ cancels the exponential factor, giving
\[
  q(z_t\mid z_{t-1};\theta_{\star})
  =\frac{c^{*}\big(z_t;\nu(t)\big)\,c^{*}\big(z_{t-1}-z_t;\Delta_t\big)}
        {c^{*}\big(z_{t-1};\nu(t-1)\big)},
\]
This recovers \cref{eq:thin}. Conditioning on the sufficient statistic $Z_{t-1}$ removes $\theta_{\star}$ for every supported exponent, including $1<p<2$.

\subsection{Gaussian, Poisson, and Gamma kernels}

For $p\in\{0,1,2\}$, the normalizing factor $c^{*}$ has an explicit form that gives a standard thinning distribution. We identify each kernel by substituting $c^{*}$ into \cref{eq:thin}. Throughout, $\alpha_t=\nu(t)/\nu(t-1)$ and $\Delta_t/\nu(t-1)=1-\alpha_t$. The results are summarized in \cref{tab:families}.

\medskip\noindent\textbf{Gaussian ($p=0$).}
Here $c^{*}(z;\nu)\propto e^{-z^{2}/2\nu\sigma^{2}}$, so the numerator is
proportional to
\[
  \exp\left\{
    -\frac{z_t^{2}}{2\nu(t)\sigma^{2}}
    -\frac{(z_{t-1}-z_t)^{2}}{2\Delta_t\sigma^{2}}
  \right\},
\]
a Gaussian in $z_t$. Collecting the quadratic and linear coefficients,
\begin{align*}
  \text{precision}
  &=\frac{1}{\sigma^{2}}\left[\frac{1}{\nu(t)}+\frac{1}{\Delta_t}\right]
   =\frac{\nu(t-1)}{\sigma^{2}\,\nu(t)\,\Delta_t},\\[4pt]
  \text{linear coefficient}
  &=\frac{z_{t-1}}{\Delta_t\sigma^{2}} .
\end{align*}
The mean is the ratio of the two and the variance the reciprocal of the
precision:
\begin{align*}
  \text{mean}
  &=\frac{z_{t-1}}{\Delta_t\sigma^{2}}\cdot
    \frac{\sigma^{2}\nu(t)\Delta_t}{\nu(t-1)}
   =\frac{\nu(t)}{\nu(t-1)}\,z_{t-1}
   =\alpha_t z_{t-1},\\[4pt]
  \text{variance}
  &=\frac{\sigma^{2}\,\nu(t)\,\Delta_t}{\nu(t-1)}
   =\sigma^{2}\alpha_t\Delta_t .
\end{align*}
Thus, $z_t\mid z_{t-1}\sim\mathcal{N}(\alpha_tz_{t-1},\,\sigma^{2}\alpha_t\Delta_t)$, with a mean proportional to the observed total and a variance determined by the schedule.

\medskip\noindent\textbf{Poisson ($p=1$).}
Here $\kappa(\theta)=e^{\theta}$ and $Z\sim\mathrm{Pois}(\nu e^{\theta})$, so
\[
  c^{*}(z;\nu)=\frac{\nu^{z}}{z!},
  \qquad z\in\mathbb{N}.
\]
Substituting,
\begin{align*}
  q(z_t\mid z_{t-1})
  &=\frac{\nu(t)^{z_t}}{z_t!}
    \cdot\frac{\Delta_t^{\,z_{t-1}-z_t}}{(z_{t-1}-z_t)!}
    \cdot\frac{z_{t-1}!}{\nu(t-1)^{z_{t-1}}}\\[4pt]
  &=\binom{z_{t-1}}{z_t}
    \left(\frac{\nu(t)}{\nu(t-1)}\right)^{\!z_t}
    \left(\frac{\Delta_t}{\nu(t-1)}\right)^{\!z_{t-1}-z_t}\\[4pt]
  &=\binom{z_{t-1}}{z_t}\,
    \alpha_t^{z_t}\,(1-\alpha_t)^{z_{t-1}-z_t},
\end{align*}
The resulting kernel is $\mathrm{Binomial}(z_{t-1},\alpha_t)$, which retains each count independently with probability $\alpha_t$.

\medskip\noindent\textbf{Gamma ($p=2$).}
Here $\kappa(\theta)=-\log(-\theta)$ and $Z\sim\Gamma(\nu,-\theta)$, so
\[
  c^{*}(z;\nu)=\frac{z^{\nu-1}}{\Gamma(\nu)} .
\]
Substituting,
\begin{align*}
  q(z_t\mid z_{t-1})
  &=\frac{\Gamma\big(\nu(t-1)\big)}
         {\Gamma\big(\nu(t)\big)\,\Gamma(\Delta_t)}\\[4pt]
  &\qquad\times
   z_t^{\nu(t)-1}\,
   (z_{t-1}-z_t)^{\Delta_t-1}\,
   z_{t-1}^{\,1-\nu(t-1)} .
\end{align*}
Change variable to $u=z_t/z_{t-1}$. The powers of $z_{t-1}$ collect as
\[
  z_{t-1}^{\,\nu(t)-1}\cdot z_{t-1}^{\,\Delta_t-1}\cdot z_{t-1}^{\,1-\nu(t-1)}
  \cdot z_{t-1}
  \;=\; z_{t-1}^{\,\nu(t)+\Delta_t-\nu(t-1)}
  \;=\; z_{t-1}^{\,0},
\]
The final factor is the Jacobian $dz_t=z_{t-1}\,du$. All dependence on $z_{t-1}$ cancels, leaving the Beta density
\[
  f_U(u)=\frac{u^{\nu(t)-1}(1-u)^{\Delta_t-1}}
    {B\big(\nu(t),\Delta_t\big)},\qquad 0<u<1.
\]
Hence $U\sim\mathrm{Beta}(\nu(t),\Delta_t)$ independently of the total $Z_{t-1}$~\cite{lukacs1955}, and thinning scales that total by $U$.

\subsection{Latent thinning for intermediate exponents}

For $1<p<2$, \cref{prop:latent} combines binomial thinning of the latent count with Beta thinning of the accumulated amplitude.

The jump count satisfies $N_\nu\sim\mathrm{Pois}(\lambda\nu)$. Applying the Poisson thinning kernel gives
\[
  N_t \mid N_{t-1}=n \;\sim\; \mathrm{Binomial}(n,\alpha_t),
  \qquad
  \alpha_t=\frac{\nu(t)}{\nu(t-1)} .
\]
Because jump amplitudes are independent of the count process, their distribution is unchanged by count thinning.

For the amplitude layer, condition on $N_t=n_1$ and $N_{t-1}=n$. The total at
$t$ is a sum of $n_1$ independent $\Gamma(\alpha,\beta)$ jumps and the
discarded part is the sum of the remaining $n-n_1$, so by Gamma additivity at
the common scale $\beta$,
\[
  Z_t\sim\Gamma(n_1\alpha,\beta),
  \qquad
  Z_{t-1}-Z_t\sim\Gamma\big((n-n_1)\alpha,\beta\big),
\]
These totals are independent and share a scale. For $0<n_1<n$, the Beta--Gamma identity~\cite{lukacs1955} gives
\[
  \frac{Z_t}{Z_t+(Z_{t-1}-Z_t)}
  =\frac{Z_t}{Z_{t-1}}
  \;\sim\;\mathrm{Beta}\big(n_1\alpha,(n-n_1)\alpha\big),
\]
independent of $Z_{t-1}$ given the counts. The scale $\beta$ cancels in both
layers, so neither depends on it.

For the boundary counts, $n_1=0$ gives $Z_t=0$, and $n_1=n$ gives $Z_t=Z_{t-1}$. When $n=0$, both totals are zero and no ratio is formed. These cases recover the atoms of the mixed thinning kernel without evaluating a Beta distribution with a zero shape parameter.

The shape $\alpha=(2-p)/(p-1)$ is fixed by the Tweedie parameterization with variance function $\mu^p$. It equals the negative of the transformed exponent in \cref{eq:s-stable} for $1<p<2$.

\subsection{Latent-count inference for real observations}
\label{sec:sup-latent}

Both thinning layers are exact conditional on $N$. Under synthetic degradation the simulator can retain the generated count alongside the state, avoiding count inference. A real observation supplies only
the total $Z$, and the count has to be sampled from its posterior. Writing the
compound density as a sum over $n$ and cancelling every factor that does not
depend on $n$, the weights are, at unit dispersion and for $z>0$,
\begin{equation}
\begin{aligned}
  \Pr[N_\nu=n\mid Z_\nu=z]&\;\propto\;\frac{A^{n}}{n!\,\Gamma(n\alpha)},\\
  A&=\frac{\nu\,z^{\alpha}}{(2-p)\,(p-1)^{\alpha}},
\end{aligned}
  \label{eq:s-latent}
\end{equation}
for $n\ge1$, with $z=0$ forcing $n=0$. The mean drops out of the $n$-dependence:
the two places $\mu$ enters, the Poisson rate $\lambda\propto\mu^{2-p}$ and the
jump scale $\beta\propto\mu^{\,p-1}$, cancel because
$\alpha(p-1)=2-p$. \cref{eq:s-latent} is the Tweedie
series~\cite{dunn2005} read as a distribution over $n$ rather than as a density
in $z$.

The implementation uses $n_{\max}=64$, normalizes these weights over $1\le n\le n_{\max}$, and samples by inverse cumulative probability. For fixed $(p,\nu,z)$, define the omitted posterior mass
\[
  \epsilon_{\mathrm{tail}}(z)=
  \frac{\sum_{n>n_{\max}} A^n/[n!\Gamma(n\alpha)]}
       {\sum_{n\ge1} A^n/[n!\Gamma(n\alpha)]}.
\]
The omitted posterior mass $\epsilon_{\mathrm{tail}}(z)$ equals the total variation error of the truncated count distribution.

Interior-exponent thinning of real observations requires this posterior step, whereas Gamma thinning uses the exact Beta kernel. Our synthetic severity-transfer protocol instead uses simulator-retained counts to thin the conditioning channel. The reported transfer results therefore apply to retained-count thinning. Transfer using inferred counts remains to be evaluated.

\subsection{Relation to the excess-noise factor}

To derive \cref{eq:excess}, let the gain of each primary event follow $G\sim\Gamma(k,\beta)$ in shape--scale form. Its excess-noise factor is
\[
  F^{2}:=\frac{\E[G^{2}]}{\E[G]^{2}} .
\]
For a Gamma gain, $\E[G]=k\beta$ and $\Var(G)=k\beta^{2}$, so
\begin{equation}
  F^{2}=1+\frac{\Var(G)}{\E[G]^{2}}
       =1+\frac{k\beta^{2}}{k^{2}\beta^{2}}
       =1+\frac{1}{k},
  \label{eq:s-fano}
\end{equation}
This factor is independent of $\beta$. Identifying the gain shape with the jump shape in \cref{prop:latent}, $\alpha=k=(2-p)/(p-1)$, gives
\[
  \frac{2-p}{p-1}=k
  \quad\Longrightarrow\quad
  2-p=k(p-1)
  \quad\Longrightarrow\quad
  p=\frac{2+k}{1+k} .
\]
Inverting \cref{eq:s-fano} as $k=1/(F^{2}-1)$ and substituting,
\[
  p=\frac{2+\dfrac{1}{F^{2}-1}}{1+\dfrac{1}{F^{2}-1}}
   =\frac{2(F^{2}-1)+1}{(F^{2}-1)+1}
   =\frac{2F^{2}-1}{F^{2}}
   =2-\frac{1}{F^{2}},
\]
which is \cref{eq:excess}. The map is strictly increasing in $F^{2}$, with
\[
  p\to1^{+}\ \text{ as }\ F^{2}\to1^{+},
  \qquad
  p\to2^{-}\ \text{ as }\ F^{2}\to\infty,
\]
hence a bijection onto $(1,2)$. Holding the mean and measurement index fixed while varying the jump rate and scale according to the Tweedie parameterization recovers the Poisson and Gamma marginals at the endpoints.

This relation applies within the Tweedie compound parameterization. With dispersion $\phi$, its parameters are
\[
  \lambda=\frac{\mu^{2-p}}{\phi(2-p)},
  \qquad
  \alpha=\frac{2-p}{p-1},
  \qquad
  \beta=\phi(p-1)\,\mu^{\,p-1} ,
\]
which reduces to \cref{eq:cpg-params} at $\phi=1$. In particular, the jump scale depends on the mean. For comparison, hold the gain parameters $\alpha$ and $\beta$ fixed and vary only the event rate. With $N\sim\mathrm{Pois}(\lambda)$, $X_i\stackrel{iid}{\sim}\Gamma(\alpha,\beta)$, and $Z=\sum_{i\le N}X_i$,
\begin{align*}
  \E[Z]&=\lambda\,\alpha\beta, \\
  \Var(Z)&=\lambda\,\E[X_1^{2}]=\lambda\,\alpha(\alpha+1)\beta^{2}, \\
  \frac{\Var(Z)}{\E[Z]}&=(\alpha+1)\,\beta ,
\end{align*}
The variance-to-mean ratio is independent of $\lambda$, giving a linear variance law with dispersion $(\alpha+1)\beta$. The distribution remains compound Poisson, rather than the scaled Poisson law of a Tweedie $p=1$ model. Thus, \cref{eq:excess} relates gain shape to the exponent under the full Tweedie construction. Assigning an exponent to a sensor also requires variance--intensity calibration, as in the FMD and SAR experiments.

\section{Output parameterization and index calibration}
\label{sec:sup-anchor}

The variance law in \cref{eq:fwd} explains the exponent dependence of relative noise and the sensitivity of the measurement index to intensity scaling.

\subsection{Relative noise and residual anchoring}

By \cref{eq:fwd}, $\E[x_t]=x_{\star}$ and $\Var(x_t)=V(x_{\star})/\nu(t)$, so with
$V(\mu)=\mu^{p}$,
\begin{align*}
  \mathrm{CV}(x_t\mid x_{\star})
  &=\frac{\sqrt{\Var(x_t)}}{\E[x_t]}
   =\frac{\sqrt{V(x_{\star})/\nu(t)}}{x_{\star}}\\[4pt]
  &=\frac{x_{\star}^{\,p/2}}{x_{\star}\sqrt{\nu(t)}}
   =\frac{x_{\star}^{\,p/2-1}}{\sqrt{\nu(t)}} ,
\end{align*}
This recovers \cref{eq:cv}. At a fixed measurement index, relative noise is signal-independent for $p=2$ and increases towards low intensities for $p<2$. This behavior motivates log-intensity prediction for $1\le p<2$ and residual anchoring at $p=2$.

\subsection{Effect of radiometric gain}

On a homogeneous patch write $\hat\nu:=V(\bar x)/\hat\sigma^{2}$ with $\bar x$
the sample mean and $\hat\sigma^{2}$ the sample variance. Under a radiometric
rescaling $x\mapsto cx$ the two sample moments transform as
\[
  \bar x\mapsto c\,\bar x,
  \qquad
  \hat\sigma^{2}\mapsto c^{2}\hat\sigma^{2},
\]
so with $V(\mu)=\mu^{p}$,
\[
  \hat\nu=\frac{V(\bar x)}{\hat\sigma^{2}}
  \;\longmapsto\;
  \frac{(c\bar x)^{p}}{c^{2}\hat\sigma^{2}}
  = c^{\,p-2}\,\frac{\bar x^{\,p}}{\hat\sigma^{2}}
  = c^{\,p-2}\,\hat\nu ,
\]
This gives \cref{eq:scale}. Gamma index estimation is gain-invariant because $p-2=0$ at $p=2$. For other exponents, the estimated index depends on the intensity units, so calibration requires a fixed normalization convention.

\subsection{Prediction domain}

The supported $p\ge1$ models have positive mean domains and nonnegative measurement support. We enforce positive predictions through the output parameterization. At $p=0$, the Gaussian mean domain is $\mathbb R$, and we use an identity map. Projection onto the bounded working range $\XW$ is a separate implementation choice.

\section{Exponent range and Gaussian special case}
\label{sec:sup-ends}

The main text studies $\{0\}\cup[1,2]$. Here we summarize the exponent domain of the Tweedie family and derive the Gaussian special case.

\subsection{Admissible exponents}

Tweedie exponential dispersion models with variance function $V(\mu)=\mu^p$ exist for $p\notin(0,1)$~\cite{jorgensen1987,jorgensen1997,dunn2005}. A commonly used exponent transformation is
\begin{equation}
  \alpha_{\mathrm{s}}=\frac{p-2}{p-1},
  \label{eq:s-stable}
\end{equation}
for $p\ne1$. In the compound Poisson--Gamma range, $-\alpha_{\mathrm{s}}=(2-p)/(p-1)$ gives the positive jump shape rather than an ordinary stable-law index. Positive-support models also exist for $p>2$, including the inverse Gaussian model at $p=3$, for which bridge simulation methods are available. These models lie outside the scope of the present framework.

\subsection{Gaussian marginals and reverse update}

At $p=0$ the variance function is the constant $\sigma^{2}$, so
\[
  \kappa(\theta)=\tfrac12\sigma^{2}\theta^{2},
  \qquad
  \EDs(\theta,\nu)=\mathcal{N}\big(\nu\sigma^{2}\theta,\ \nu\sigma^{2}\big).
\]
Since $\tau(\theta)=\kappa'(\theta)=\sigma^{2}\theta$, the bridge's natural
parameter is
\[
  \theta_{\star}=\tau^{-1}(x_{\star})=\frac{x_{\star}}{\sigma^{2}},
\]
so that, writing $\sigma_t^{2}:=\sigma^{2}/\nu(t)$,
\[
  Z_t\sim\mathcal{N}\big(\nu(t)x_{\star},\ \nu(t)\sigma^{2}\big)
  \quad\Longrightarrow\quad
  x_t=\frac{Z_t}{\nu(t)}\sim\mathcal{N}\big(x_{\star},\ \sigma_t^{2}\big),
\]
This gives the Gaussian reparameterization $x_t=x_{\star}+\sigma_t\xi$ in \cref{eq:gauss}, with $\xi\sim\mathcal{N}(0,I)$. The conditional variance is independent of the clean intensity, and the shared reverse-mean update in \cref{eq:reverse} applies directly.

Although the marginal parameterization is Gaussian, the deterministic update differs from a probability-flow ODE step. Since $\alpha_t=\nu(t)/\nu(t-1)=\sigma_{t-1}^{2}/\sigma_t^{2}$, \cref{eq:reverse} gives
\[
  x_{t-1}=\hat x_{\star}+\frac{\sigma_{t-1}^{2}}{\sigma_t^{2}}\,(x_t-\hat x_{\star}),
\]
whereas an Euler step on the EDM probability-flow ODE in the same
$\sigma$ parameterisation~\cite{karras2022} gives
\[
  x_{t-1}=\hat x_{\star}+\frac{\sigma_{t-1}}{\sigma_t}\,(x_t-\hat x_{\star}),
\]
The coefficients are therefore the variance ratio and standard-deviation ratio, respectively. For example, a step from $\sigma=2$ to $\sigma=1$ retains one quarter of the displacement under our update and one half under the Euler update. Our coefficient follows from the conditional mean of the superposition coupling in \cref{thm:reverse}.

\section{Role of the consistency objective}
\label{sec:sup-cons}

The reconstruction and consistency objectives both compare predictions with $x_{\star}$, but evaluate the network on different state distributions. Reconstruction uses states from the forward process, whereas consistency uses states generated by a model update.

Suppose the network is pointwise optimal, meaning
\[
  \hat x_{\star}(x_t;t,\xobs)=\E[x_{\star}\mid x_t,\xobs]
  \quad\text{at every }(x_t,t)\text{ visited,}
\]
so that $\nabla_\theta\mathcal{L}_{\mathrm{rec}}=0$ there. Applying the chain
rule to \cref{eq:l-cons}, with the unified gradient \cref{eq:dp-grad} supplying
the inner derivative,
\begin{equation}
  \nabla_\theta\mathcal{L}_{\mathrm{cons}}
  =\E\!\left[
     -\frac{2\,\nu(t')\,\big(x_{\star}-\hat x_{\star}(x_{t'})\big)}
            {V\big(\hat x_{\star}(x_{t'})\big)}\,
     \nabla_\theta\hat x_{\star}(x_{t'})
   \right].
  \label{eq:s-consgrad}
\end{equation}
The stop-gradient removes derivatives through the generated state $x_{t'}$, while the network is still evaluated on that state. The induced joint distribution of $(x_{\star},x_{t'},\xobs)$ can differ from the forward training distribution at $t'$. Its optimal conditional-mean predictor can therefore differ as well, so optimal reconstruction on forward states does not imply that \cref{eq:s-consgrad} vanishes. The consistency term addresses this possible distribution shift.

\section{Evaluation conventions}
\label{sec:sup-conv}

We provide additional results on the sampling budget and worst-case evaluation.

\begin{table}[t]
  \centering
  \scriptsize
  \caption{Effect of the number of function evaluations at severe noise. ``Diagonal'' denotes matched model and data exponents. ``Worst'' is the minimum over the tested data exponents. Bold marks the preferred value in each row.}
  \label{tab:s-nfe}
  \begin{tabular}{lccc}
    \toprule
    severe & $\NFE{=}1$ & $\NFE{=}5$ & $\NFE{=}25$ \\
    \midrule
    diagonal mean PSNR         & 22.326 & \textbf{22.330} & 22.225 \\
    $p{=}1.5$ worst over $\pd$ & 20.739 & 21.339 & \textbf{21.594} \\
    mean off-diagonal penalty  & $+1.601$ & $+1.256$ & $\mathbf{+0.924}$ \\
    diagonal dark log-RMSE     & 0.2841 & \textbf{0.2831} & 0.2854 \\
    diagonal res.\ var.        & 0.926 & 0.949 & \textbf{0.968} \\
    \bottomrule
  \end{tabular}
\end{table}

\subsection{Number of function evaluations}
\label{sec:sup-nfe}
\Cref{tab:s-nfe} evaluates fixed-exponent models with one, five, and $25$ network evaluations at severe noise. For matched exponents, five evaluations achieve a mean PSNR of $22.330\dB$, compared with $22.326\dB$ for one evaluation and $22.225\dB$ for $25$. Dark-region log-RMSE is also lowest at five evaluations ($0.2831$), compared with $0.2841$ and $0.2854$ at one and $25$ evaluations, respectively. Additional evaluations are more helpful under exponent mismatch. From one to $25$ evaluations, the minimum PSNR of the fixed $p=1.5$ model across tested data exponents increases by $0.855\dB$, while the mean mismatch penalty decreases from $1.601$ to $0.924\dB$. Thus, longer sampling reduces mismatch losses but does not improve matched-exponent PSNR in this comparison.

We further evaluate the conditional model with $\NFE\in\{1,3,5,8,10,15,25\}$ on a held-out split. PSNR peaks at three evaluations and decreases slightly thereafter. Relative to this peak, one, five, and $25$ evaluations reduce PSNR by $0.021$, $0.006$, and $0.092\dB$, respectively. All runs use deterministic sampling, and varying the projection onto $\XW$ changes PSNR by at most $0.002\dB$. We retain five evaluations as the common sampling budget, with a gap of only $0.006\dB$ to the best setting in this sweep.

\subsection{Evaluation across noise exponents}
\label{sec:sup-minimax}
Mean mismatch loss and worst-case PSNR describe different aspects of exponent robustness. The former summarizes the cost of mismatched model--data exponent pairs and depends on the evaluation grid. At severe noise with five network evaluations, it is $2.000\dB$ on the $3\times3$ grid and $1.256\dB$ on the $5\times5$ grid. The latter is the minimum PSNR across the tested data exponents. We use this common criterion in \cref{tab:cond} to compare fixed-exponent models with the conditional network. Among the fixed models, the minimax model maximizes this minimum.

\section{Additional experimental results}
\label{sec:sup-tables}

\subsection{Additional restoration examples}

\begin{figure*}[p]
  \centering
  \includegraphics[width=0.75\textwidth]{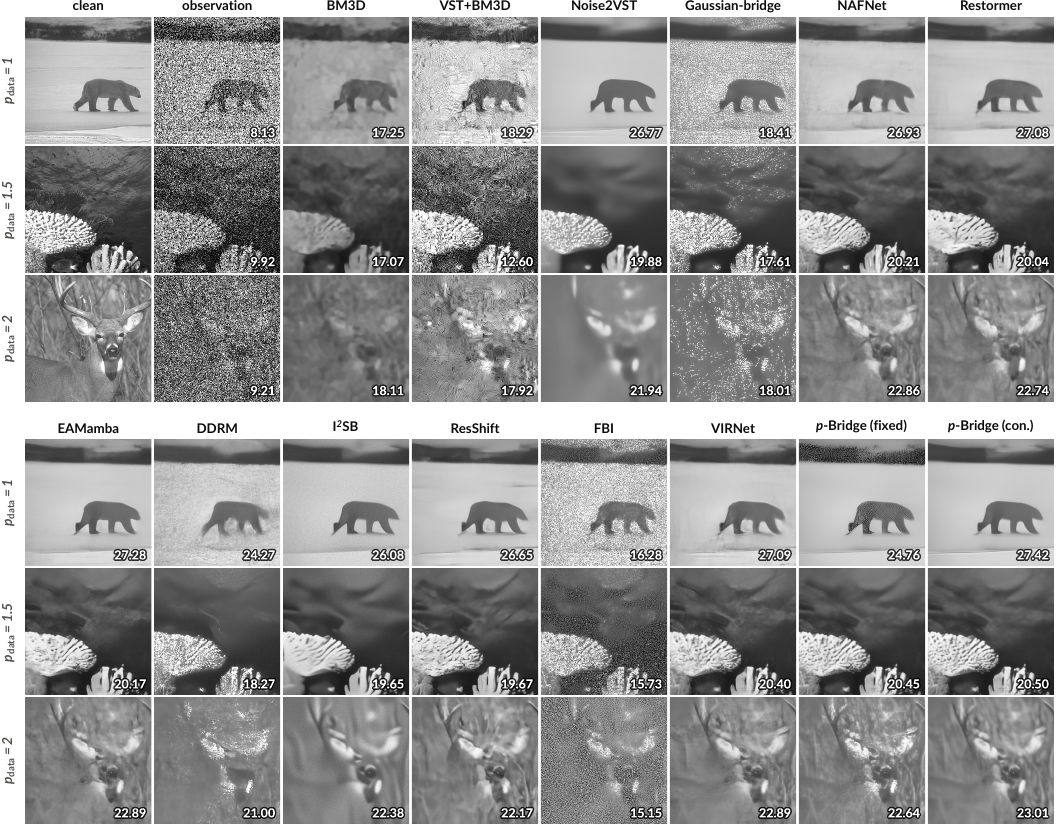}
  \caption{Synthetic-image restorations at severe noise ($\eta=1$), with one row per noise exponent. Corner values show PSNR (dB).}
  \label{fig:grid-severe}
\end{figure*}

\begin{figure*}[p]
  \centering
  \includegraphics[width=0.75\textwidth]{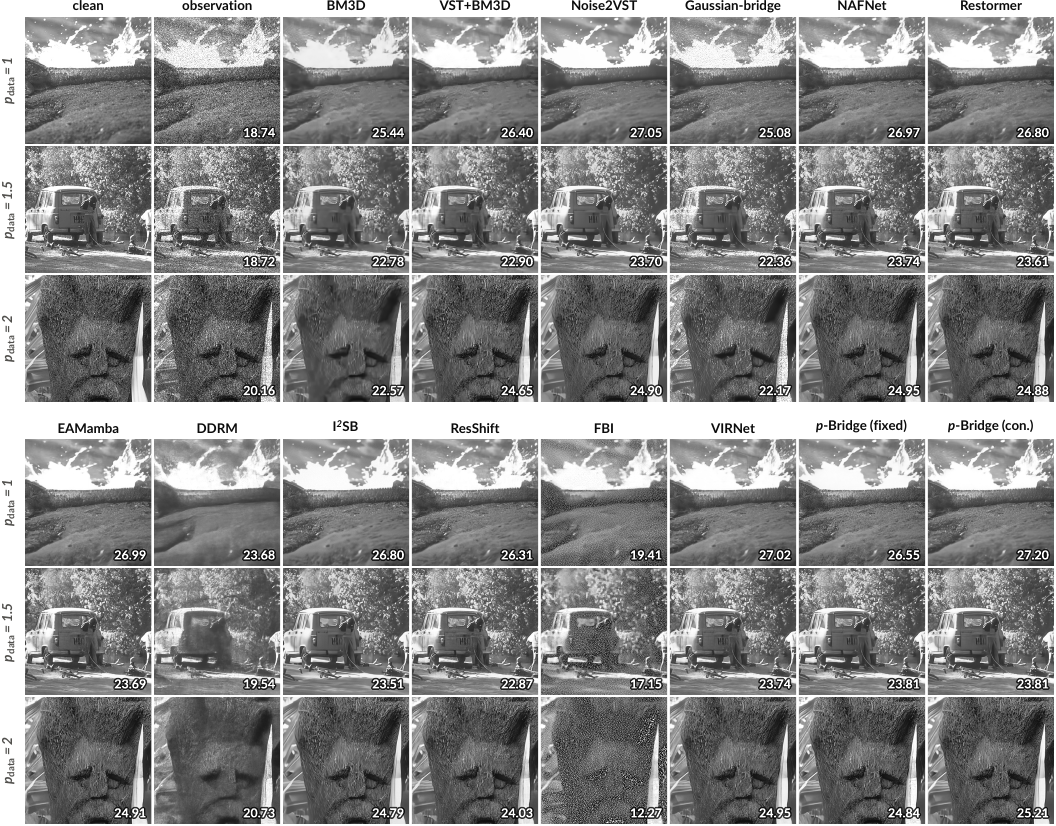}
  \caption{Synthetic-image restorations at mild noise ($\eta=0.25$), with one row per noise exponent. Corner values show PSNR (dB).}
  \label{fig:grid-mild}
\end{figure*}

\Cref{fig:grid-severe,fig:grid-mild} show synthetic-image restorations across noise exponents at severe and mild noise, complementing the moderate-noise comparison in \cref{fig:grid-moderate}. Corner values report PSNR for each displayed crop.

\subsection{Sensitivity to estimator settings}
\label{sec:sup-estimator-robustness}
We assess the sensitivity of noise-exponent estimation to three settings: local window sizes $\{9,15,25\}$, intensity-bin counts $\{6,10,16\}$, and retained fractions $\{0.1,0.3,0.5\}$ of the flattest samples within each bin. Across the tested settings, the maximum absolute exponent error $|\hat p-p|$ ranges from $0.067$ to $0.089$. On constant-intensity tiles, $|\hat p-p|\leq0.0066$, and $\nu$ is recovered to three significant figures.

\section{Implementation details}
\label{sec:sup-impl}

\subsection{Conditioning and training-step sampling}
\label{sec:obj-arch}
The observation is supplied through a separate conditioning channel at the training measurement index $\nuobs$. This input remains fixed throughout sampling and is adjusted separately for noise-severity transfer.

We sample training steps uniformly from $\{t_{\min},\ldots,T-1\}$. The expectation in \cref{eq:l-rec} is therefore evaluated as
\[
  w=\frac{1}{T-t_{\min}}
       \sum_{t=t_{\min}}^{T-1}\frac{\nu(t)}{\nuobs}.
\]
\subsection{The conditional model}
\label{sec:obj-cond}
We cache the $21$ bridge configurations corresponding to the training exponents. Each optimizer step accumulates gradients over $12$ micro-batches, each with its own sampled exponent.

\phantomsection\label{sec:exp-rebalance}
\noindent\textbf{Exponent-dependent loss weighting.} During conditional training, we multiply the reconstruction deviance by $\muref^{\,2-p}$ as an empirical weighting across exponents. \Cref{tab:ablation} evaluates this choice by replacing the factor with $1$.

At inference, we update the exponent embedding and bridge configuration together, including the measurement index corresponding to the selected exponent.

\subsection{Noise-severity transfer}
\label{sec:obj-transfer}

For noise-severity transfer, only the conditioning input is thinned to the training index $\nuobs$, while the observed image used to initialize sampling is unchanged. For $1<p<2$, synthetic evaluations use simulator-retained latent counts for this thinning step.

\subsection{Noise-exponent estimation}
\label{sec:obj-phat}
\begin{algorithm}[t]
  \caption{Single-image noise-parameter estimation}
  \label{alg:noise-estimation}
  \footnotesize
  \begin{algorithmic}[1]
    \Require Noisy intensity image $x$; $h=15$, $B=10$, $\rho=0.3$
    \Ensure Regression estimates $(\hat p,\hat\nu)$ under $\phi=1$
    \State $r_{i,j}\gets\tfrac12(x_{i,j}-x_{i+1,j}-x_{i,j+1}+x_{i+1,j+1})$
    \State $\bar x\gets\Call{BoxMean}{x,h\times h}$
    \State Partition samples into $B$ bins by $\bar x$
    \For{each intensity bin $b$}
      \State Retain the flattest fraction $\rho$ of samples
      \State Estimate bin mean $m_b$ and noise variance $v_b$
      \Statex \hspace{\algorithmicindent}\hspace{\algorithmicindent}from the retained local means and residuals
    \EndFor
    \State $(\hat p,a)\gets\Call{TheilSen}{\{(\log m_b,\log v_b)\}_{b=1}^{B}}$
    \State $\hat\nu\gets\exp(-a)$
    \State \Return $(\hat p,\hat\nu)$
  \end{algorithmic}
\end{algorithm}

\begin{table}[t]
	\centering\scriptsize
	\caption{SAR noise-exponent estimates across local-ENL window selections (medians over images). Nominal and dispersion-implied looks, $L$, are included for reference.}
	\label{tab:s-sar-p}
	\setlength{\tabcolsep}{3.2pt}
	\begin{tabular}{lcccccccc}
		\toprule
		\multirow{2}{*}{sensor} & nominal &
		\multicolumn{5}{c}{$\hat p$ at ENL band} & implied \\
		\cmidrule(lr){3-7}
		& $L$ & all & 5--95 & 10--95 & 25--95 & 40--95 & $L$ \\
		\midrule
		miniSAR & 1.00 & 2.150 & 2.042 & 2.014 & 2.005 & 2.002 & 1.06 \\
		FARAD-X & 1.00 & 2.103 & 2.071 & 2.058 & 2.030 & 2.013 & 0.92 \\
		TerraSAR-X & 1.16 & 2.106 & 2.067 & 2.044 & 2.014 & 2.005 & 1.37 \\
		\bottomrule
	\end{tabular}
\end{table}

\Cref{alg:noise-estimation} summarizes the single-image estimator. The cross-difference residual removes locally affine signals. Under independent pixel noise, its variance equals the average noise variance of the four pixels. A $15\times15$ box filter supplies the local mean intensities used for binning. Retaining the flattest $30\%$ within each of $10$ bins limits the contribution of image structure to the variance estimates, which are then used in the Theil--Sen fit. The pseudocode returns the regression estimates before the range clipping described in the main paper.

\subsection{Baseline configurations}
\label{sec:sup-base}

I$^2$SB and ResShift use our UNet and training pool. ResShift operates in pixel space with five stochastic sampling steps. DDRM uses an unconditional DDPM prior trained on the clean pool. On synthetic data, its scalar noise level is computed from the true per-pixel variance. Noise2VST fits a transform to each test image for $2{,}000$ iterations with frozen pretrained denoisers. Its reported $\NFE=1$ counts the final denoising pass. EAMamba uses $128\times128$ training crops, while evaluation uses $256\times256$ crops.

\section{SAR data and supplementary evaluation}
\label{sec:sup-sar}

We convert Sandia miniSAR/FARAD complex data and ESA TerraSAR-X 16-bit MGD-SE amplitudes to linear intensity, excluding display-rendered or terrain-resampled images. \Cref{tab:s-sar-p} shows local-ENL sensitivity: at the 40--95 percentiles, fitted slopes span $2.002$--$2.016$, close to the Gamma branch's quadratic variance law.

\end{document}